\documentclass[lettersize,journal]{IEEEtran}
\usepackage{amsmath,amsfonts}
\usepackage{array}
\usepackage{textcomp}
\usepackage{stfloats}
\usepackage{url}
\usepackage{verbatim}
\usepackage{graphicx}
\usepackage{cite}
\usepackage{amsfonts,amsmath,amssymb,mathtools}
\usepackage{algorithm}
\usepackage[noend]{algpseudocode}
\usepackage{subcaption}
\usepackage{multirow}
\usepackage{graphicx}
\usepackage{wrapfig,lipsum}
\usepackage{xcolor}         
\usepackage{paralist}
\usepackage{tabularx}

\definecolor{LightCyan}{rgb}{0.88,1,1}

\def\vdelta{{\boldsymbol \delta}}

\def\bw{{\boldsymbol w}}

\DeclareMathOperator*{\argmin}{argmin}

\renewcommand{\phi}{\varphi}

\def\bs{{\boldsymbol s}}

\def\bw{{\boldsymbol w}}
\def\bx{{\boldsymbol x}}

\def\bmu{{\boldsymbol \mu}}

\newcommand{\E}{\mathbb{E}}

\newtheorem{prop}{Proposition}

\newcommand{\argmax}{\operatornamewithlimits{argmax}}
\newcommand\numberthis{\addtocounter{equation}{1}\tag{\theequation}}

\begin{document}

\title{Adversarial Examples Detection with Bayesian Neural Network}

\author{%
  \IEEEauthorblockN{%
    Yao Li, 
    Tongyi Tang, 
    Cho-Jui Hsieh and  
    Thomas C. M. Lee,  
    ~\IEEEmembership{Senior Member,~IEEE}
  } 
\thanks{This work was supported in part by the National Science Foundation under grants CCF-1934568, IIS-2048280, IIS-2008173, DMS-2113605, DMS-2210388, DMS-2152289, DMS-2134107, and Cisco Faculty Award.}
\thanks{Yao Li is with the Statistics and Operations Research Department, University of North Carolina at Chapel Hill, NC. (e-mail: yaoli@email.unc.edu).}
\thanks{Tongyi Tang is with the Statistics Department, University of California Davis, CA. (e-mail: tyitang@ucdavis.edu).}
\thanks{Cho-Jui Hsieh is with the Computer Science Department, University of California Los Angeles, CA. (e-mail: chohsieh@cs.ucla.edu).}
\thanks{Thomas C.M. Lee is with the Statistics Department, University of California Davis, CA. (e-mail:tcmlee@ucdavis.edu)}
\thanks{\copyright 2024 IEEE. Personal use of this material is permitted. Permission from IEEE must be obtained for all other uses, in any current or future media, including reprinting/republishing this material for advertising or promotional purposes, creating new collective works, for resale or redistribution to servers or lists, or reuse of any copyrighted component of this work in other works.}}



\maketitle


\begin{abstract}
In this paper, we propose a new framework to detect adversarial examples motivated by the observations that random components can improve the smoothness of predictors and make it easier to simulate the output distribution of a deep neural network. With these observations, we propose a novel Bayesian adversarial example detector, short for \textsc{BATer}, to improve the performance of adversarial example detection. Specifically, we study the distributional difference of hidden layer output between natural and adversarial examples, and propose to use the randomness of the Bayesian neural network to simulate hidden layer output distribution and leverage the distribution dispersion to detect adversarial examples. 
The advantage of a Bayesian neural network is that the output is stochastic while a deep neural network without random components does not have such characteristics. 
Empirical results on several benchmark datasets against popular attacks show that the proposed \textsc{BATer} outperforms the state-of-the-art detectors in adversarial example detection.
\end{abstract}

\begin{IEEEkeywords}
adversarial example, deep neural network, Bayesian neural network, detection
\end{IEEEkeywords}

\section{Introduction}
\label{sec:intro}

Despite achieving tremendous successes, Deep Neural Networks (DNNs) have been shown to be vulnerable against adversarial attacks~\cite{goodfellow2014explaining,szegedy2013intriguing,kuurkova2018artificial,yang2020towards,li2020robustness,li2022review}. By adding imperceptible perturbations to the original inputs, the attackers can craft adversarial examples to fool a trained classifier. Adversarial examples are indistinguishable from the original inputs to humans but are mis-classified by the classifier. The wide application of machine learning models causes concerns about the reliability and safety of machine learning systems in security-sensitive areas, such as self-driving, financial systems, and healthcare. 

There has been extensive research on improving the robustness of deep neural networks against adversarial examples~\cite{yuan2019attack,zhang2020oppo,chan2021break,li2021review}. In~\cite{athalye2018obfuscated}, the authors showed that many defense methods~\cite{dhillon2018stochastic,ma2018characterizing,samangouei2018defense,song2017pixeldefend,xie2017mitigating} can be circumvented by strong attacks except Madry's adversarial training~\cite{madry2017towards}, in which adversarial examples are generated during training and added back to the training set. Since then, adversarial training-based algorithms have become state-of-the-art methods for defending against adversarial examples. However, despite being able to improve robustness under strong attacks, adversarial training-based algorithms are time-consuming due to the cost of generating adversarial examples on-the-fly. Improving the robustness of deep neural networks remains an open question.

Due to the difficulty of defense, recent work has turned to attempting to detect adversarial examples as an alternative solution. The main assumption made by the detectors is that adversarial samples come from a distribution that is different from the natural data distribution, that is, adversarial samples do not lie on the data manifold, and DNNs perform correctly only near the manifold of the training data~\cite{tanay2016boundary}. Many works have been done to study the characteristics of adversarial examples and leverage the characteristics to detect adversarial examples instead of trying to classify them correctly~\cite{ma2018characterizing,feinman2017detecting,zheng2018robust,pang2018towards,tao2018attacks,lee2018simple,yang2020ml,ding2022consensus}.

\begin{figure*}
    \centering
    \includegraphics[width=0.95\textwidth]{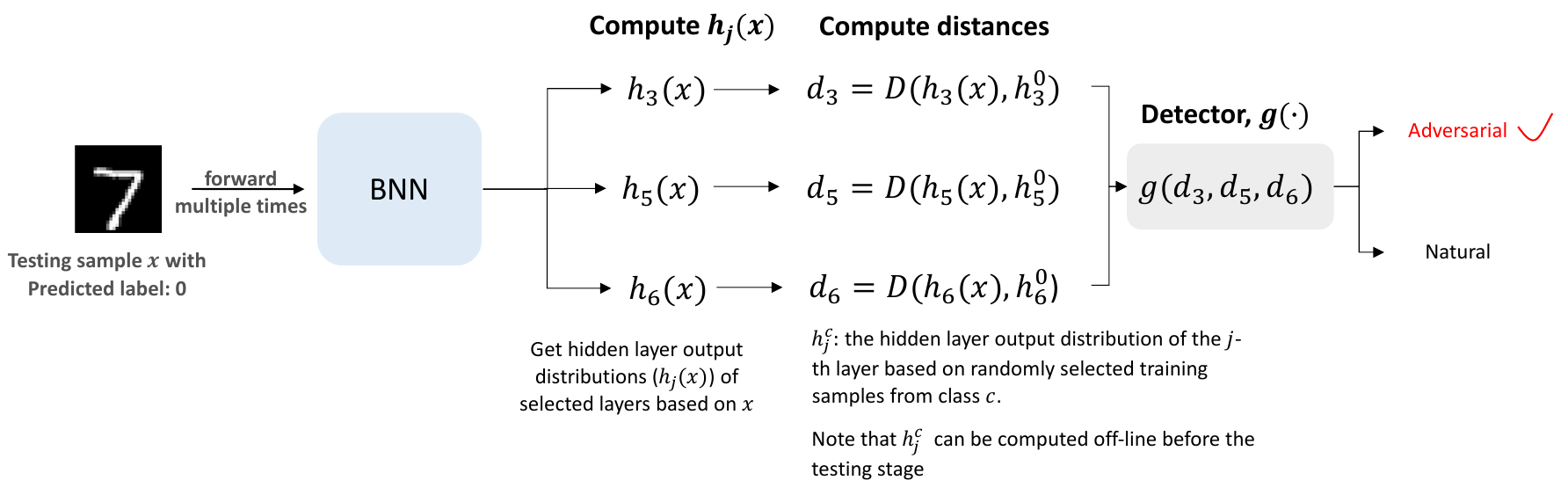}
    \vspace{-5pt}
    \caption{Detection framework of \textsc{BATer}. An example is given in this diagram to show how \textsc{BATer} works. An adversarial image ($\bx$) with handwritten digit 7 is mis-classified as 0 by the classifier (BNN). To check if the input is adversarial or not, the input image is fed into the BNN multiple times to get hidden layer output distributions ($h_j(\bx)$) of selected layers (layers 3, 5, and 6 in this example). Details of $h_j(\bx)$ computation and layer selection are given in Section~\ref{sec:method}. Then, distances ($d_j$) between hidden layer output distributions ($h_j(\bx)$) and hidden layer output distributions based on training samples of predicted class ($h_j^c$) are computed. In this example, it is $h_j^0$ because the model predicts the input as class 0. Finally, the distances are fed into the detector to do binary classification: adversarial vs. natural. Details of distance computation and detector training can be found in Section~\ref{sec:method}.
    }
    \label{fig:bd_frame}
    \vspace{-0.3cm}
\end{figure*}

\IEEEpubidadjcol

Despite many algorithms that have been proposed for adversarial detection, most of them are deterministic, which means they can only use the information from one single forward pass to detect adversarial examples. This makes it easier for an attacker to break those models, especially when the attacker knows the neural network architecture and weights. In this paper, we propose a novel algorithm to detect adversarial examples based on randomized neural networks. Intuitively, incorporating randomness in neural networks can improve the smoothness of predictors, thus enabling stronger robustness guarantees (see randomized-based defense methods in~\cite{liu2017towards,xie2017mitigating,cohen2019certified}). Further,
 instead of observing only one hidden feature for each layer, a randomized network can lead to a distribution of hidden features, making it easier to detect an out-of-manifold example. 

\paragraph{Contribution and Novelty} We propose a detection method based on Bayesian Neural Network (BNN), leveraging the randomness of BNN to improve detection performance (see the framework in Figure~\ref{fig:bd_frame}). BNN and some other random components have been used to improve robust classification accuracy~\cite{liu2017towards,xie2017mitigating,liu2018advbnn,ye2018bal,cohen2019certified,carbone2020bnn,yang2022mitigating}, but they were not used to improve adversarial detection performance. The proposed method \textsc{BATer} is motivated by the following observations: 1) the hidden layer output generated from adversarial examples demonstrates different characteristics from that generated from natural data and this phenomenon is more obvious in BNN than in deterministic deep neural networks; 2) randomness of BNN makes it easier to simulate the characteristics of hidden layer output. Training BNN is not very time-consuming as it only doubles the number of parameters of the deep neural network with the same structure~\cite{blundell2015weight}. However, BNN can achieve comparable classification accuracy and improve the smoothness of the classifier. A theoretical analysis is provided to show the advantage of BNN over DNN in adversarial detection.

In numerical experiments, our method achieves better performance in detecting adversarial examples generated from popular attack methods on MNIST, CIFAR10 and ImageNet-Sub among state-of-the-art detection methods. Ablation experiments show that BNN performs better than deterministic neural networks under the same detection scheme. {Besides, the proposed method is also tested against attacks with different parameters, transfer attacks, and an adaptive attack. In all the tested scenarios, the proposed method can achieve reasonable performance.}

\paragraph{Notation} In this paper, all the vectors are represented as bold symbols. The input to the classifier is represented by $\bx$ and the label associated with the input is represented by $y$. Thus, one observation is a pair $(\bx,y)$. The classifier is denoted as $f(\cdot)$ and $f(\bx)$ represents the output vector of the classifier. $f(\bx)_i$ is the score of predicting $\bx$ with label $i$. The prediction of the classifier is denoted as $c(\bx)=\argmax\limits_if(\bx)_i$; that is, the predicted label is the one with the highest prediction score. We use the $\ell_\infty$ and $\ell_2$ distortion metrics to measure similarity and report the $\ell_\infty$ distance in the normalized $[0,1]$ space (e.g., a distortion of $0.031$ corresponds to $8/256$), and the $\ell_2$ distance as the total root-mean-square distortion normalized by the total number of pixels~\cite{li2022review}. 

\section{Related Work}
\label{sec:related}

\paragraph{Adversarial attack} Multiple attack methods have been introduced for crafting adversarial examples to attack deep neural networks~\cite{yuan2019adversarial,athalye2018obfuscated,carlini2017adversarial,carlini2017towards,moosavi2017universal,che2020smgea,liang2021explore,zhao2019admm,chen2019adversarial}. Depending on the information available to the adversary, attack methods can be divided into white-box attacks and black-box attacks. Under the white-box setting, the adversary is allowed to analytically compute the model's gradients/parameters, and has full access to the model architecture. Most white-box attacks generate adversarial examples based on the gradient of the loss function with respect to the input~\cite{moosavi2016deepfool,chen2017ead,madry2017towards,carlini2017towards,carlini2018evaluation,chen2018ead}. Among them FGSM~\cite{goodfellow2014explaining}, {C$\&$W~\cite{carlini2017towards}} and PGD~\cite{madry2017towards} attacks have been widely used to test the robustness of machine learning models. In reality, the detailed model information, such as the gradient, may not be available to the attackers~\cite{li2022review}. Some attack methods are more agnostic and only rely on the predicted labels or scores~\cite{chen2017zoo,brendel2017decision, ilyas2018black, cheng2019sign, yan2019subspace}. In~\cite{chen2017zoo}, the authors proposed a method to estimate the gradient based on the score information and craft adversarial examples with the estimated gradient. Some other works~\cite{brendel2017decision, ilyas2018black, cheng2019sign, yan2019subspace,chen2019hopskipjumpattack,chen2019towards} introduced methods that also only rely on the final decision of the model.

\paragraph{Adversarial defense} To defend against adversarial examples, many studies have been done to improve the robustness of deep neural networks, including adversarial training~\cite{madry2017towards,kurakin2016adversarial,tramer2017ensemble,zhang2019theoretically,ye2021acce}, generative models~\cite{samangouei2018defense,meng2017magnet,li2020optimal,jalal2017robust,li2021towards}, verifiable defense~\cite{wong2018provable,everett2021cert} and other techniques~\cite{chen2021feature,liu2021model,zhang2020challenging,mustafa2020deeply,zhao2022enhanced,ding2022consensus}. The authors of \cite{athalye2018obfuscated} showed that many defense methods~\cite{dhillon2018stochastic,ma2018characterizing,samangouei2018defense,song2017pixeldefend,xie2017mitigating} could be circumvented by strong attacks except Madry's adversarial training~\cite{madry2017towards}. Since then, adversarial training-based algorithms have become state-of-the-art methods in defending against adversarial examples. However, adversarial training is computationally expensive and time-consuming due to the cost of generating adversarial examples on-the-fly, thus adversarial defense is still an open problem to solve.

\paragraph{Adversarial detection} Another popular line of research focuses on screening out adversarial examples~\cite{metzen2017on,metzen2017detecting,agarwal2021damad,nesti2021detect,ding2022consensus}. 
A straightforward way towards adversarial example detection is to build a simple binary classifier separating the adversarial apart from the clean data based on the characteristics of adversarial examples~\cite{metzen2017on,gong2017adversarial,feinman2017detecting,lee2018simple,sperl2020dla,gao2023detecting,chen2022adversarial}. In~\cite{ding2022consensus}, a detection method is implemented based on the consensus of the classifications of the augmented examples, which are generated based on an individually implemented intensity exchange on the red, green, and blue components of the input image. 
In~\cite{feinman2017detecting}, the author proposed to perform kernel density estimation on the training data in the feature space of the last hidden layer to help detect adversarial examples (KD). The authors of~\cite{ma2018characterizing} observed that the Local Intrinsic Dimensions (LID) of hidden-layer outputs differ between the original inputs and adversarial examples, and leveraged these findings to detect adversarial examples. 
In~\cite{lee2018simple}, an adversarial detection method based on Mahalanobis distance (MAHA) is proposed. Class conditional Gaussian distributions are first fitted based on the hidden layer output features of the deep neural network, then confidence scores are calculated to compute Mahalanobis distance.
In~\cite{yang2020ml}, the author studied the feature attributions of adversarial examples and proposed a detection method (ML-LOO) based on feature attribution scores. The author of~\cite{roth2019odds} showed that adversarial examples exist in cone-like regions in very specific directions from their corresponding natural inputs and proposed a new test statistic to detect adversarial examples with the findings (ODD). Recently, a joint statistical test pooling information from multiple layers is proposed in~\cite{raghuram2020detecting} to detect adversarial examples (JTLA). 
We show that \textsc{BATer} performs comparable or superior to these detection methods across multiple benchmark datasets. 

Recently, there has been a shift in focus towards detecting adversarial examples that are generated using black-box methods~\cite{gao2023towards}, which are recognized as more realistic threats. Despite being well explored in the vision domain, adversarial example detection started to get attention in the field of natural language processing (NLP) recently~\cite{zhou-etal-2019-learning, mozes-etal-2021-frequency, yoo-etal-2022-detection, yin2022addmu}. In addition to the domain of NLP, adversarial detection has been extended to the physical world, aiming to identify adversarial examples in real-world scenarios~\cite{ren2021adversarial}.

\begin{figure}
    \centering
     \includegraphics[width=0.4\textwidth]{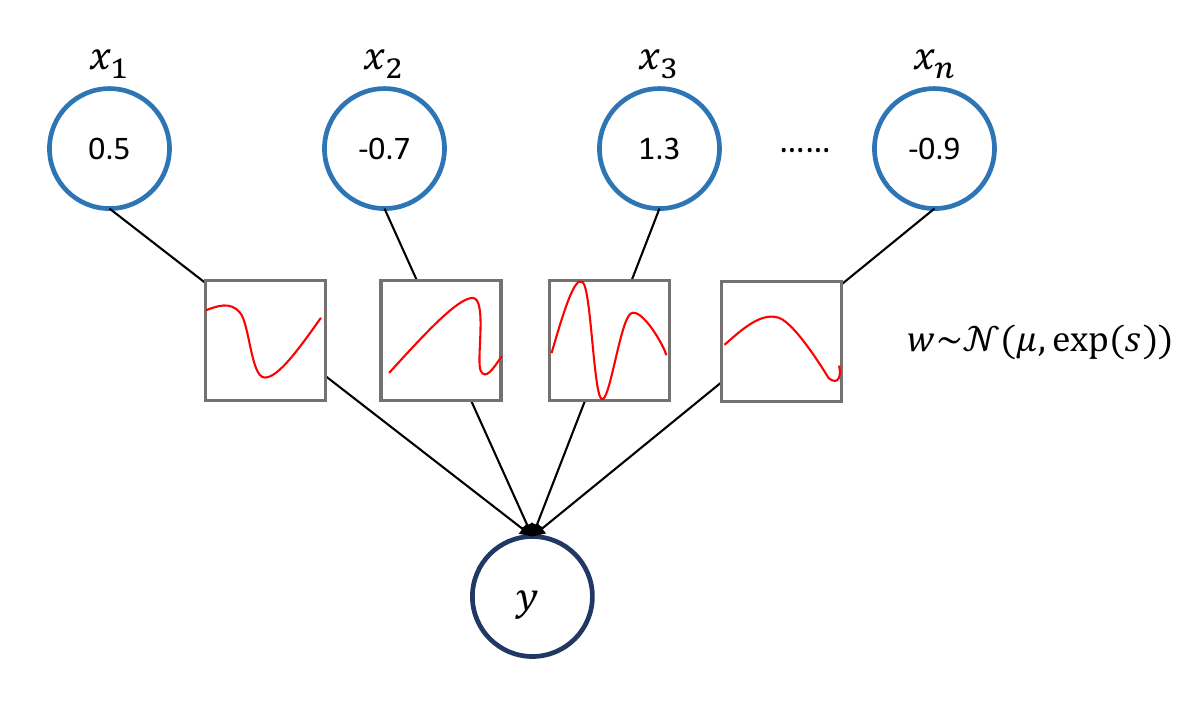}
     \vspace{-5pt}
    \caption{ Illustration of Bayesian Neural Network. {All weights in a BNN are represented by probability distributions over possible values, rather than having a single fixed value. The red curves in the graph represent distributions. We view a BNN as a probabilistic model: given an input $\bx$, a BNN assigns a probability to each possible output $y$, using the set of parameters $\bw$ sampled from the learned distributions.} }
    \label{fig:bnn}
    \vspace{-10pt}
\end{figure}

\paragraph{Bayesian neural network} The idea of BNN is illustrated in Figure~\ref{fig:bnn}. In~\cite{blundell2015weight}, the author introduced an efficient algorithm to learn the parameters of BNN. Given the observable random variables $(\bx, y)$, BNN aims to estimate the distributions of hidden variables $\bw$, instead of estimating the maximum likelihood value $\bw_{\mathrm{MLE}}$ for the weights. Since, in the Bayesian perspective, each parameter is now a random variable measuring the uncertainty of the estimation, the model can potentially extract more information to support a better prediction (in terms of precision, robustness, etc.). 

Given the input $\bx$ and label $y$, a BNN aims to estimate the posterior over the weights $p(\bw|\bx, y)$ given the prior $p(\bw)$. The true posterior can be approximated by a parametric distribution $q_{\boldsymbol{\theta}}(\bw)$, where the unknown parameter $\boldsymbol{\theta}$ is estimated by minimizing the KL divergence
\begin{equation}
    \mathsf{KL}\big(q_{\boldsymbol{\theta}}(\bw)\ \|\ p(\bw|\bx, y)\big)
    \label{eq:bnn_0}
\end{equation}
over $\boldsymbol{\theta}$. For simplicity, $q_{\boldsymbol{\theta}}$ is often assumed to be a fully factorized Gaussian distribution:
\begin{equation}
    \label{eq:mean-field}
    q_{\boldsymbol{\theta}}(\bw)=\prod_{i=1}^{d} q_{\boldsymbol{\theta}_i}(\bw_i), \text{ and } q_{\boldsymbol{\theta}_i}(\bw_i)=\mathcal{N}(\bw_i; \boldsymbol{\mu}_i, \exp(\bs_i)^2),
\end{equation}
where $\bmu$ and $\bs$ are parameters of the Gaussian distributions of weight. The objective function for training BNN is reformulated from expression~\eqref{eq:bnn_0} and is shown in expression~\eqref{eq:bnn_1}, which is a sum of a data-dependent part and a regularization part:
\begin{align*}
    \mathop{\arg\max}_{\bmu,\bs}\Big\{\sum_{(\bx_i,y_i)\in\boldsymbol{D}}\E_{\bw\sim q_{\bmu,\bs}}\log p(y_i|\bx_i,\bw)\\
    -\mathsf{KL}\big(q_{\bmu,\bs}(\bw)~\|~p(\bw)\big)\Big\},
    \numberthis \label{eq:bnn_1}
\end{align*}
where $\boldsymbol{D}$ represents the data distribution. In the first term of objective~\eqref{eq:bnn_1}, the probability of $y_i$ given $\bx_i$ and weights is the output of the model. This part represents the classification loss. The second term of objective~\eqref{eq:bnn_1} is trying to minimize the divergence between the prior and the parametric distribution, which can be viewed as regularization~\cite{blundell2015weight}. The author of~\cite{carbone2020bnn} showed that the posterior average of the gradients of BNN makes it more robust than DNN against gradient-based adversarial attacks. Though the idea of using BNN to improve robustness against adversarial examples is not new~\cite{liu2018advbnn,ye2018bal}, the previous works did not leverage BNN to help detect adversarial examples. In~\cite{liu2018advbnn,ye2018bal}, BNN was combined with adversarial training~\cite{madry2017towards} to improve robust classification accuracy. 

\section{Proposed Method}
\label{sec:method}

We first discuss the motivation behind the proposed method: 1) the distributions of the hidden layer neurons of a deep neural network can be different when based on adversarial examples versus natural images; 2) this dispersion is more obvious in BNN than DNN; 
3) it is easier to simulate hidden layer output distribution with random components. Then, we introduce the specific metric used to measure this distributional difference and extend the detection method to multiple layers to make it more resistant to adversarial attacks.

\subsection{Motivation: distributional difference of natural and adversarial hidden layer outputs}

Given input $\bx$ and a classifier $f(\cdot)$, the prediction of the classifier is denoted as $c(\bx)=\argmax\limits_i f(\bx)_i$; that is, the predicted label is the one with the highest prediction score.
The adversary aims to  perturb the original input to change the predicted label:
\begin{align*}
    c(\bx)\ne\argmax\limits_i f(\bx+\vdelta)_i,
\end{align*}
where $\vdelta$ denotes the perturbation added to the original input. The attacker aims to find a small $\vdelta$ (usually lies within a small $\ell_p$ norm ball) to successfully change the prediction of the model. 
Thus, given the same predicted label, there could be a distributional difference in hidden layer outputs between adversarial examples and natural data. For example, adversarial examples mis-classified as airplanes could have hidden layer output distributions different from those of natural airplane images. 
Here, we define a hidden layer output distribution in DNN as the empirical distribution of all the neuron values of that layer, which means all output values of that layer will be used to draw an one-dimensional histogram to simulate the hidden layer distribution. For BNN, a similar approach is used to estimate the hidden layer output distribution.
Meanwhile, in BNN, the same input will be forwarded multiple times as the weights of BNN are stochastic to get a better estimation of the output distribution. 

\begin{figure*}
  \begin{subfigure}{0.32\textwidth}
    \includegraphics[width=\linewidth]{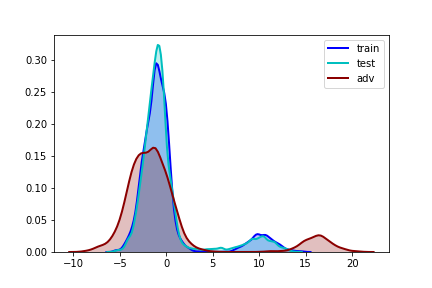}
    \caption{BNN: Layer 23} \label{fig:1a}
  \end{subfigure}%
  \hspace*{\fill}   
  \begin{subfigure}{0.32\textwidth}
    \includegraphics[width=\linewidth]{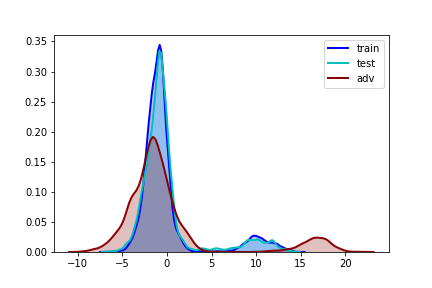}
    \caption{BNN: Layer 33} \label{fig:1b}
  \end{subfigure}%
  \hspace*{\fill}   
  \begin{subfigure}{0.32\textwidth}
    \includegraphics[width=\linewidth]{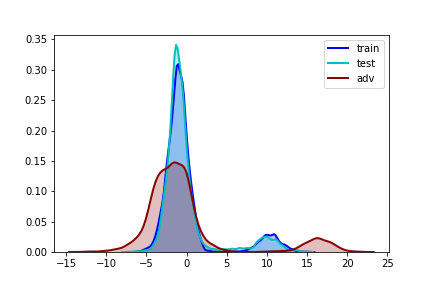}
    \caption{BNN: Layer 43} \label{fig:1c}
  \end{subfigure}

  \begin{subfigure}{0.32\textwidth}
    \includegraphics[width=\linewidth]{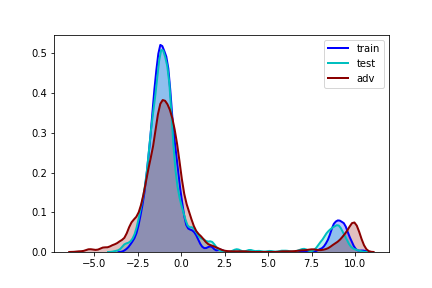}
    \caption{DNN: Layer 23} \label{fig:1d}
  \end{subfigure}%
  \hspace*{\fill}   
  \begin{subfigure}{0.32\textwidth}
    \includegraphics[width=\linewidth]{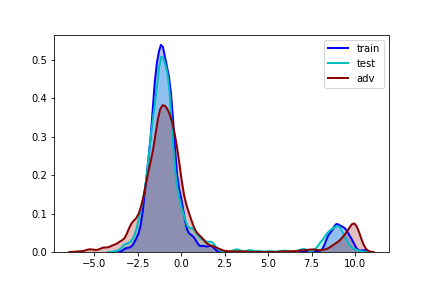}
    \caption{DNN: Layer 33} \label{fig:1e}
  \end{subfigure}%
  \hspace*{\fill}   
  \begin{subfigure}{0.32\textwidth}
    \includegraphics[width=\linewidth]{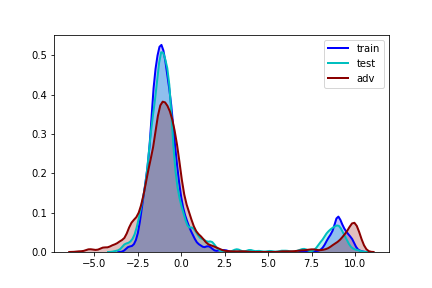}
    \caption{DNN: Layer 43} \label{fig:1f}
  \end{subfigure}
\caption{Hidden Layer output Distributions (HLDs) of VGG16 and BNN (VGG16 based architecture) based on images from automobile class of CIFAR10. Legend explanation: {\color{blue}train} represents HLDs of training samples from automobile class; {\color{cyan}test} denotes HLDs of testing samples from automobile class; {\color{red}adv} shows HLDs of adversarial examples predicted as automobiles. The adversarial examples are generated by PGD~\cite{madry2017towards}. The three plots in the first row show hidden layer distributions of a BNN, and the plots in the second row show the distributions of a DNN with the same base architecture. For both DNN and BNN, there are distributional differences between natural ({\color{blue}train} and {\color{cyan}test}) and adversarial ({\color{red}adv}) hidden outputs, but the differences are larger for BNN.} 
\label{fig:bd_dis}
\vspace{-0.3cm}
\end{figure*}

In the exploratory analysis, we compare the hidden layer output distributions of DNN and BNN based on both natural and adversarial examples, and find some interesting patterns that are later used in the proposed method. Some examples of hidden layer output distribution comparisons are shown in Figure~\ref{fig:bd_dis}.
The figure shows the hidden layer output distributions of layer $23$, layer $33$ and layer $43$ in DNN and BNN. Blue and cyan (train and test) curves represent distributions of the natural automobile images in CIFAR10. Red curves represent the distributions of adversarial examples mis-classified as automobiles. The adversarial examples are generated by PGD~\cite{madry2017towards} with $\ell_\infty$ norm. The architecture of the DNN is VGG16~\cite{simonyan2014very} and the architecture of the BNN is also VGG16~\cite{simonyan2014very} except that the weights in BNN follow Gaussian distributions. Both networks are trained on CIFAR10 train set.

In Figure~\ref{fig:bd_dis}, we can see that for all three hidden layers, there are differences between distributions based on natural and adversarial images. In BNN, the hidden layer output distributions of the natural images (train or test) are clearly different from those of adversarial examples (adv), while the pattern is not that obvious in DNN. Even though hidden layer output distributions of only three layers are shown here, similar patterns are observed in some other layers in BNN. This phenomenon is not a special case with PGD adversarial examples on CIFAR10. Such characteristics are also found in adversarial examples generated by different attack methods on other datasets. 

\paragraph{Why BNN not DNN?} Differences between distributions based on natural and adversarial examples can be observed in both DNN and BNN. However, the distributional difference is more obvious in BNN than in neural networks without random components (see Figure~\ref{fig:bd_dis}). Therefore, more information can be extracted from BNN than from deterministic neural networks. Furthermore, random components of BNN make it easier to simulate the hidden layer output distributions. Our experimental results also show that the proposed detection method works better with BNN than with deterministic neural networks on multiple datasets (see Section~\ref{sec:ablation} for more details).

Figure~\ref{fig:bd_dis} empirically shows the intuition behind the proposed framework. The following theoretical analysis shows that randomness can help enlarge the distributional differences between natural and adversarial hidden layer outputs. 
\begin{prop}
Let $f(\bx, \bw)$ be a model with $\bx\sim \boldsymbol{D}_\bx$ and $\bw \sim \boldsymbol{D}_\bw$, where $\boldsymbol{D}_\bw$ is any distribution that satisfies $\bw$ is symmetric about $\bw_0 = \mathbb{E}[\bw]$, such as $\mathcal{N}(\bw_0,\boldsymbol{I})$. If $
\nabla_\bx f(\bx, \bw)$ can be approximated by the first order Taylor expansion at $\bw_0$, we have 
\begin{align}
  \mathcal{D}(f(\bx+\vdelta, \bw), f(\bx, \bw)) \geq \mathcal{D}(f(\bx+\vdelta, \bw_0), f(\bx, \bw_0)),  
\end{align}
where $\vdelta$ represents adversarial perturbation and $\mathcal{D}$ represents a translation-invariant distance measuring distribution dispersion (See proof of the inequality in Appendix~\ref{sec:theo}).
\end{prop}
The inequality shows that randomness involved in parameters will enlarge the distributional differences between natural and adversarial outputs. Therefore, leveraging the hidden layer output distributional differences of BNN to detect adversarial examples is a sensible choice.

\subsection{Detect adversarial examples by distribution distance}
\label{sec: detect_frame}

We propose to measure the dispersion between hidden layer output distributions of adversarial examples and natural inputs and use this characteristic to detect adversarial examples. In particular, given an input $\bx$ and its predicted label $c$, we measure the distribution distance between the hidden layer output distribution of $\bx$ and the corresponding hidden layer output distribution of training samples from class $c$:
\begin{align}
    d_j(\bx) = \mathcal{D}\left(h_j(\bx), h_j(\{\bx^c_i\}_{i=1}^{n_c})\right),
\end{align}
where $h_j(\bx)$ represents the hidden layer output distribution of the $j$-th layer based on testing sample $\bx$, $h_j(\{\bx^c_i\}_{i=1}^{n_c})$ represents the hidden layer output distribution of the $j$-th layer based on training samples from class $c$, $n_c$ is the number of training samples in class $c$, and $\mathcal{D}$ can be arbitrary divergence. 
{For simplicity, $h_j(\{\bx^c_i\}_{i=1}^{n_c})$ is replaced by $h_j^c$ in the rest part of the paper. Besides, $n_c$ does not have to be the total number of training samples in class $c$. In our experiments, $n_c$ is just a small amount sampled from the training samples of class $c$. }
As for the measure of divergence, we estimate the divergence with 1-Wasserstein distance in our experiments. However, other divergence measures can also be used, such as the Kullback–Leibler divergence. 

The hidden layer output distribution is estimated by a one-dimensional empirical distribution of all the output values of that layer. 
The hidden layer output distribution ($h_j^c$) estimated with training samples of each class can be easily simulated since there are multiple samples in each class. 
However, at the testing stage, only one testing sample ($\bx$) is available for the simulation of $h_j(\bx)$.
For a deep neural network without random components, the hidden layer output is deterministic, thus the simulation result depends on a single forward pass. For BNN, the hidden layer output is stochastic, thus we can simulate the distribution with multiple passes. 

To pool the information from different levels, the dispersion is measured at multiple hidden layers to generate a set of dispersion scores $\{d_j|j\in\mathcal{S}\}$, where $\mathcal{S}$ is the index set of selected hidden layers (see details of layer selection in Section~\ref{sec: imp_detail}). It is expected that natural inputs will have small dispersion scores while adversarial examples will have relatively large dispersion scores. A binary classifier is trained on the dispersion scores to detect adversarial examples. In the paper, we fit a binomial logistic regression model to do the binary classification. An overview of the detection framework at testing time is shown in Figure~\ref{fig:bd_frame}. Details of the method are included in Algorithm~\ref{alg:bd}.

\setlength{\textfloatsep}{5pt}
\begin{algorithm}
  
\caption{\textsc{BATer}}\label{alg:bd}
\textbf{Input:} Input $\bx$, pre-trained BNN $f(\cdot)$, pre-trained binary classifier $g(\cdot)$, number of passes to simulate hidden layer output distribution $B$, indices of hidden layers selected for detection $\mathcal{S}$ and divergence $\mathcal{D}$. \\
\textbf{Output:} Adversarial ($z=1$) or Natural ($z=0$).
\begin{algorithmic}[1]
\State $c = \argmax\limits_if(\bx)_i$
            \Comment{\emph{get the predicted label $c$}}
\For {$j\in\mathcal{S}$}
\State Feed $\bx$ into $f(\cdot)$ $B$ times to simulate $h_j(\bx)$
\State $d_j = \mathcal{D}(h_j(\bx), h_j^c)$ 
    \Comment{\emph{$h_j^c$ is the $j$-th layer output distribution of class $c$}}
\EndFor
\State $z=g(d_1, d_2, ..., d_k)$
        \Comment{\emph{$z=1$ indicating adversarial example and $z=0$ indicating natural input}}
\end{algorithmic}
\end{algorithm}

\subsection{Implementation Details}
\label{sec: imp_detail}

\paragraph{Layer Selection} For adversarial examples generated with different attacks on different datasets, 
the pattern of distributional differences can be different. For example, adversarial examples generated by PGD on CIFAR10 show larger distributional dispersion in deeper layers (layers closer to the final layer). However, such characteristic does not appear in adversarial examples generated by C$\&$W on CIFAR10. Instead, the distributional dispersion is more obvious in some front layers (layers closer to the input layer). 
Therefore, we develop an automated hidden layer selection scheme to find the layers with large deviations between natural data and adversarial examples. 
Cross-validation is performed to do layer selection by fitting a binary classifier (logistic regression) with a single layer's dispersion score. Layers with top-ranked performance measured by AUC (Area Under the receiver operating characteristic Curve) scores are selected, and information from those layers is pooled for ultimate detection (See details of selected layers in Appendix~\ref{sec:layer}).

\paragraph{Distance Calculation} To measure the dispersion between hidden layer output distributions of natural and adversarial samples, we treat the output of a hidden layer as a realization of a one-dimensional random variable. 
The dispersion between two distributions is estimated by 1-Wasserstein distance between their empirical distributions. In BNN, the empirical distribution of a testing sample can be simulated by multiple forward passes. Whereas, in DNN, a single forward pass is done to simulate the empirical distribution as the output is deterministic. Training samples from the same class can be used to simulate empirical hidden layer output distributions of natural data of that class. 
Given a testing sample and its predicted label, calculating the dispersion score with all training samples in the predicted class is expensive, so we sample some natural images in the predicted class as representatives to speed up the process.

\paragraph{Dimension Reduction} To further improve computational efficiency, we apply dimension reduction on the hidden layer output. PCA (Principal Component Analysis) is applied to the hidden layer output of training samples to do dimension reduction before the testing stage. At the testing stage, hidden layer output is projected to a lower dimension before calculating dispersion scores, which speeds up the dispersion score calculation with high-dimensional output.

\section{Experimental Results}
\label{sec:exp}

We evaluate \textsc{BATer} on the following well-known image classification datasets: MNIST~\cite{lecun1998mnist}, CIFAR10~\cite{krizhevsky2009learning} and Imagenet-sub~\cite{miyato2018spectral}. 
The training sets provided by the datasets are used to train BNN and DNN. The BNN and DNN architectures are the same, except that the weights of BNN follow Gaussian distributions. We train BNN with Gaussian variational inference because it is straightforward to implement. We have also tried to train BNN with other techniques, such as K-FAC~\cite{zhang2018noisy}, but they all generate similar results. 

The test sets are split into $20\%$ in training folds and $80\%$ in test folds. The detection models (binary classifiers) of KD, LID and \textsc{BATer} are trained on the training folds and the test folds are used to evaluate the performance of different detection methods. Foolbox~\cite{rauber2017foolbox} is used to generate adversarial examples with the following attack methods: FGSM~\cite{goodfellow2014explaining} with $\ell_\infty$ norm, PGD~\cite{madry2017towards} with $\ell_\infty$ norm and C$\&$W~\cite{carlini2017towards} with $\ell_2$ norm. Since BNN is stochastic, original PGD and C$\&$W attacks without considering randomness are not strong enough against it. For fair comparison, we update PGD and C$\&$W with stochastic optimization methods (multiple forward passes are used to estimate gradient not just one pass). 

{Experiments in Section~\ref{sec:sota} to \ref{sec:attack_param} are done in a gray-box setting, in which we assume the adversary has access to the classifier model but does not know the detector. 
An adaptive attack is proposed in Section~\ref{sec:custom_attack} to attack \textsc{BATer} in a white-box setting, in which we assume the adversary has access to both the classifier and the detector. Details of parameter selection, neural network architectures, implementation, code github and examples of detected adversarial examples are provided in the Appendix.}

\vspace{-10pt}
\subsection{Comparison with State-of-the-Art Methods}
\label{sec:sota}

We compare the performance of \textsc{BATer} with the following state-of-the-art detection methods for adversarial detection: 1) Kernel Density Detection (KD)~\cite{feinman2017detecting}, 2) Local Intrinsic Dimensionality detection (LID)~\cite{ma2018characterizing}, 3) Odds are Odd Detection (ODD)~\cite{roth2019odds}, 4) Joint statistical Testing across DNN Layers for Anomalies (JTLA)~\cite{raghuram2020detecting}. In~\cite{raghuram2020detecting}, JTLA outperforms deep Mahalanobis detection~\cite{lee2018simple}, deep KNN~\cite{papernot2018deep}, and trust score~\cite{jiang2018trust}, so we do not include the performance of the three here. Details of implementation and parameters can be found in the Appendix. All the detection methods are tested by the following attacks: 1) FGSM~\cite{goodfellow2014explaining} with $\ell_\infty$ norm bounded by 0.3, 0.03 and 0.01 for MNIST, CIFAR10 and Imagenet-sub respectively; 2) PGD~\cite{madry2017towards} with $\ell_\infty$ norm bounded by 0.3, 0.03 and 0.01 for MNIST, CIFAR10 and Imagenet-sub respectively; C$\&$W~\cite{carlini2017towards} with confidence of 0 for all three datasets.

\begin{table*}
    \centering
\resizebox{\textwidth}{!}{
    \begin{tabular}{c|c|c|c|c|c|c||c|c|c|c|c||c|c|c|c|c}
    \hline
    \multirow{2}{*}{Data}    & \multirow{2}{*}{Metric} & \multicolumn{5}{c||}{C$\&$W} &  \multicolumn{5}{c||}{FGSM}  & \multicolumn{5}{c}{PGD}  \\ \cline{3-17}
              & &  KD   &  LID  &  ODD & JTLA &  \textsc{BATer} &  KD   &  LID  &  ODD & JTLA &  \textsc{BATer} &  KD   &  LID  &  ODD & JTLA &  \textsc{BATer}  \\ \hline \hline
    \multirow{4}{*}{CIFAR10} & AUC   & 0.945  & 0.947 &  0.955 &  0.968 &  {\bf 0.980} & 0.873 & 0.957  &  0.968 & 0.990 & {\bf 0.995} & 0.791 & 0.777 &  0.963 & 0.962 & {\bf 0.971} \\ \cline{2-17}
               & TPR(FPR@0.01) & 0.068 & 0.220 & 0.591 & 0.309& {\bf 0.606} & 0.136 & 0.385& 0.224 & 0.698 & {\bf 0.878} & 0.018 & 0.093 & 0.059 & 0.191 & {\bf 0.813}\\ \cline{2-17}
               & TPR(FPR@0.05) & 0.464 & 0.668 & 0.839 & 0.726 & {\bf 0.881} & 0.401 & 0.753 & 0.709 & 0.974 & {\bf 0.991} & 0.148 & 0.317 & 0.819 & 0.789 & {\bf 0.881} \\ \cline{2-17}
               & TPR(FPR@0.10) & 0.911 & 0.856 & 0.901 & 0.954 & {\bf 0.965} & 0.572 & 0.875 & {\bf 1.000} & {\bf 1.000} & 0.998 & 0.285 & 0.448 & {\bf 0.999} & {\bf 0.999} & 0.917 \\ \hline \hline
    \multirow{4}{*}{MNIST}   & AUC     & 0.932  & 0.785 &  0.968 & 0.980 &  {\bf 0.999} & 0.933 & 0.888 &  0.952 & 0.992 &  {\bf 0.999} & 0.801 & 0.861 &  0.967 & 0.975 & {\bf 0.989} \\ \cline{2-17}
               & TPR(FPR@0.01) & 0.196 & 0.079 & 0.212 & 0.630 & {\bf 0.974}  & 0.421 & 0.152 & 0.898 & 0.885 & {\bf 0.972} & 0.062 & 0.170 & 0.607 & 0.382 & {\bf 0.733} \\ \cline{2-17}
               & TPR(FPR@0.05) & 0.616 & 0.263 & 0.911 & 0.900 & {\bf 0.997} & 0.692 & 0.503 & 0.908 & 0.990 & {\bf 0.998} & 0.275 & 0.396 & 0.934 & 0.851 & {\bf 0.957} \\ \cline{2-17}
               & TPR(FPR@0.10) & 0.818 & 0.397 & {\bf 1.000} & 0.972 & {\bf 1.000} & 0.796 & 0.678 & 0.917 & {\bf 1.000} & {\bf 1.000} & 0.429 & 0.552 & 0.945 & 0.956 & {\bf 0.999} \\ \hline \hline
\multirow{2}{*}{Imagenet}& AUC  & 0.811  & 0.905 & 0.886 & 0.834 &  {\bf 0.941} & 0.914 & 0.983 & 0.844 & 0.842 &  {\bf 0.989} & 0.989 & {\bf 0.991} & 0.777  & 0.824 & 0.976 \\ \cline{2-17}
 & TPR(FPR@0.01) & 0.193 & {\bf 0.401} & 0.185 & 0.035 & 0.146 & 0.460 & {\bf 0.772} & 0.042 & 0.045 & 0.569 & {\bf 0.930} & 0.829 & 0.010 & 0.028 & 0.729 \\ \cline{2-17}
 \multirow{2}{*}{-sub}  & TPR(FPR@0.05) & 0.452 & {\bf 0.653} & 0.398 & 0.167 & 0.538 & 0.727 & 0.952 & 0.188 & 0.197 & {\bf 0.989} & {\bf 0.966} & 0.961 & 0.054 & 0.139 & 0.904 \\ \cline{2-17}
               & TPR(FPR@0.10) & 0.584 & 0.754 & 0.566 & 0.312 & {\bf 0.815} & 0.822 & 0.987 & 0.364 & 0.358 & {\bf 1.000} & 0.979 & {\bf 0.984} & 0.121 & 0.280 & 0.947 \\ \hline
    \end{tabular}
    }
    \caption{Performance of detection methods against adversarial attacks. {The best performance among the five detection methods is marked in {\bf bold}. In general, \textsc{BATer} performs the best or comparable to the best in most cases.}}
    \label{tab:compare1}
\end{table*}

\begin{figure*}
\centering
\includegraphics[width=0.32\textwidth]{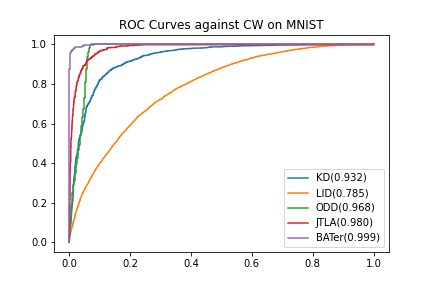}
\includegraphics[width=0.32\textwidth]{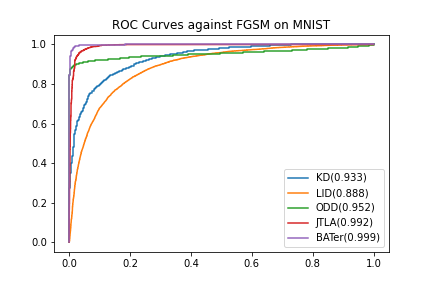}
\includegraphics[width=0.32\textwidth]{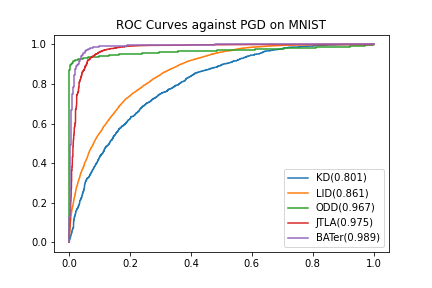}
\includegraphics[width=0.32\textwidth]{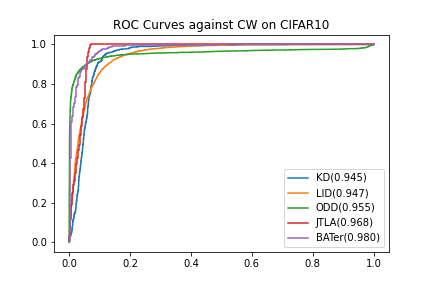}
\includegraphics[width=0.32\textwidth]{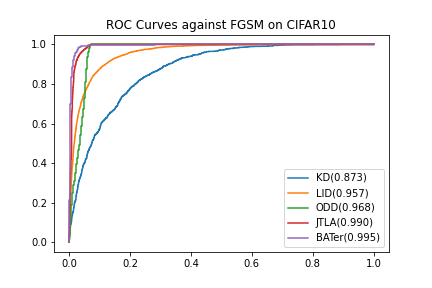}
\includegraphics[width=0.32\textwidth]{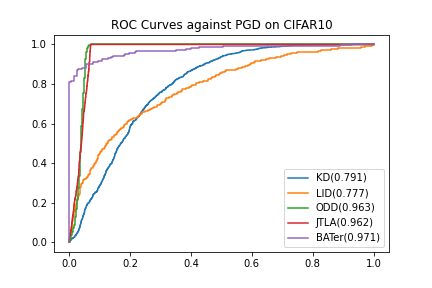}
\caption{ROC Curves of experiments in Section~\ref{sec:sota} on MNIST and CIFAR10. {The curves show that \textsc{BATer} outperforms other detection methods or perform comparably to the best method in all the cases.}} \label{fig:roc}
\end{figure*}

We report the AUC (Area Under the receiver operating characteristic Curve) score as the performance evaluation criterion as well as the True Positive Rates (TPR) by thresholding False Positive Rates (FPR) at 0.01, 0.05 and 0.1, as it is practical to keep mis-classified natural data at a low proportion. 
TPR represents the proportion of adversarial examples classified as adversarial, and FPR represents the proportion of natural data mis-classified as adversarial. 
Before calculating performance metrics, all the adversarial examples that can be classified correctly by the model are removed. The results are reported in Table~\ref{tab:compare1} and ROC curves are shown in Figure~\ref{fig:roc}. \textsc{BATer} shows superior or comparable performance over the other four detection methods across three datasets against three attacks.

\vspace{-10pt}
\subsection{Ablation Study: BNN versus DNN}
\label{sec:ablation}

\begin{figure}
\centering
    \begin{subfigure}{0.23\textwidth}
    \includegraphics[width=\textwidth]{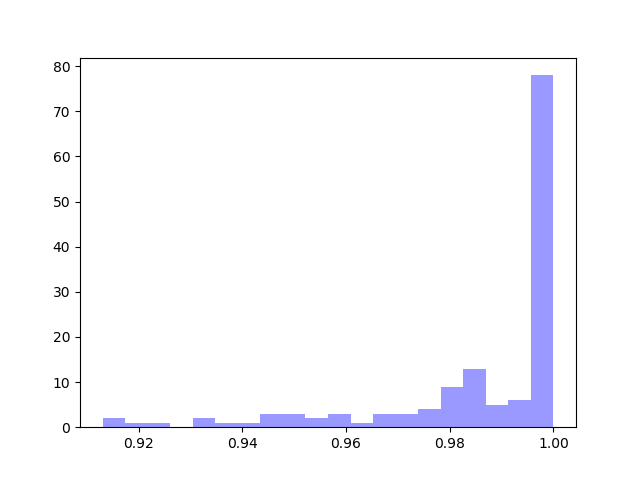}
    \caption{AUC of BNN} \label{fig:1aa}      
    \end{subfigure}
    \begin{subfigure}{0.23\textwidth}
    \includegraphics[width=\textwidth]{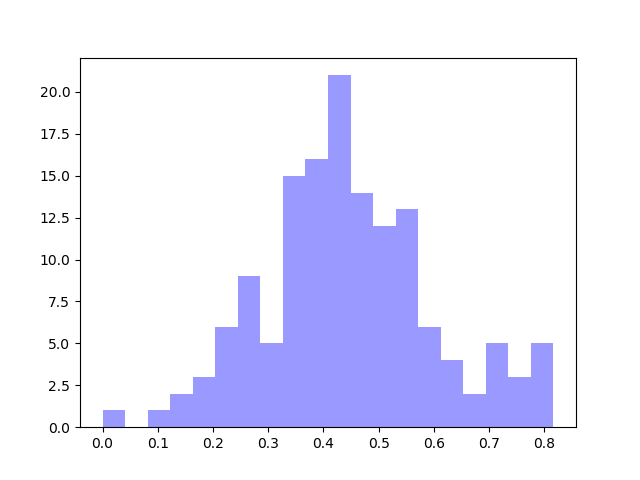}
    \caption{AUC of DNN} \label{fig:1bb}      
    \end{subfigure}
    \caption{ AUC Histograms of \textsc{BATer} with different structures (BNN vs. DNN) on Imagenet-sub. {It is obvious that BNN results in better AUCs.}}
    \label{fig:bnnvsnn}
\end{figure}

In this section, we compare the performance of \textsc{BATer} using different structures (BNN versus DNN) against PGD across three datasets. The $\ell_\infty$ norm is bounded by 0.3, 0.03 and 0.01 for MNIST, CIFAR10 and Imagenet-sub respectively. The detection methods are the same (as described in Algorithm~\ref{alg:bd}) and the differences are: 1) \textsc{BATer} with DNN uses a pre-trained deep neural network of the same structure without random components; 2) The number of passes is one as DNN does not produce different outputs with the same input. We report the class conditional AUC of the two different structures across three datasets. 

\begin{table}
    \centering
    \resizebox{0.26\textwidth}{!}{
    \begin{tabular}{c|cc|cc} \hline
    \multirow{2}{*}{Class}        &   \multicolumn{2}{c|}{CIFAR10} &   \multicolumn{2}{c}{MNIST} \\   \cline{2-5}
            &  BNN  &  DNN   &  BNN &  DNN    \\ \hline \hline
    class1  & 0.978 & 0.489 & 0.929  & 0.901    \\
    class2  & 0.972 & 0.410 & 1.000  & 0.967     \\
    class3  & 0.973 & 0.501 & 0.993  & 0.892      \\
    class4  & 0.994 & 0.594 & 0.991  & 0.958      \\
    class5  & 0.955 & 0.477 & 1.000  & 0.883      \\
    class6  & 0.995 & 0.729 & 0.999  & 0.937    \\
    class7  & 0.976 & 0.584 & 0.989  & 0.878      \\
    class8  & 0.973 & 0.537 & 1.000  & 0.941      \\
    class9  & 0.915 & 0.493 & 0.959  & 0.874    \\
    lass10  & 0.949 & 0.567 & 0.982  & 0.917     \\ \hline
    \end{tabular}
    }
    \caption{ AUCs of \textsc{BATer} with different structures (BNN vs. DNN) on CIFAR10 and MNIST of different classes. {Since in all the cases, BNN gives better results, it is clear that BNN is a better choice than DNN, which shows that random components can help improve detection performance.}}
    \label{tab:bnnvsnn}
\end{table} 

The comparison results on CIFAR10 and MNIST are shown in Table~\ref{tab:bnnvsnn} and the results on Imagenet-sub are shown in Figure~\ref{fig:bnnvsnn}. Since there are $143$ classes in Imagenet-sub, it is not reasonable to show the results in a table. Instead, we show the AUC histograms of \textsc{BATer} with different structures in Figure~\ref{fig:bnnvsnn}. Comparing the AUCs of applying \textsc{BATer} with BNN and DNN on CIFAR10 and MNIST, it is obvious that the BNN structure demonstrates superior performance all the time. On Imagenet-sub, the AUC histogram of \textsc{BATer} with BNN ranges from $0.90$ to $1.00$ and is left-tailed, while the AUC histogram of \textsc{BATer} with DNN ranges from $0.10$ to $0.85$ and centers around $0.40$, so the BNN structure clearly outperforms on Imagenet-sub. The experimental results show that random components can help improve detection results.

\vspace{-10pt}
\subsection{Transfer Attack}
\label{sec:transfer}

In this section, we study the performance of \textsc{BATer} under transfer attack setting. In practice, the defense method does not know what attack methods will be used. Therefore, defense methods trained with adversarial examples generated from one attack method may be attacked by adversarial examples generated by another attack method. When generating adversarial examples, we employ the same attack parameters as outlined in Section~\ref{sec:sota}.The performance of \textsc{BATer} in the transfer attack setting are shown in Table~\ref{tab:trans}. The results show that \textsc{BATer} trained on one type of adversarial examples can generalize to other types. 

\begin{table}
    \centering
\resizebox{0.48\textwidth}{!}{
    \begin{tabular}{c|c|c|c||c|c|c||c|c|c}
    \hline
    \multirow{2}{*}{Data}    & \multicolumn{3}{c||}{MNIST} & \multicolumn{3}{c||}{CIFAR10} &  \multicolumn{3}{c}{Imagenet-sub}  \\ \cline{2-10}
               & C$\&$W & PGD & FGSM & C$\&$W & PGD & FGSM & C$\&$W & PGD & FGSM  \\ \hline \hline
    C$\&$W & 0.999 & 0.994 & 0.994 & 0.980 & 0.877 & 0.972 & 0.941 & 0.845 & 0.870 \\
    PGD    & 0.989 & 0.989 & 0.989 & 0.820 & 0.971 & 0.824 & 0.886 & 0.976 & 0.975 \\
    FGSM   & 0.998 & 0.998 & 0.999 & 0.868 & 0.912 & 0.995 & 0.914 & 0.896 & 0.989 \\ \hline
    \end{tabular}
    }
    \caption{Performance of \textsc{BATer} under transfer attack. The column names represent the adversarial examples the detector trained with. The row names represent the adversarial examples the detector tested against. AUC scores are reported.}
    \label{tab:trans}
\end{table}

\vspace{-10pt}
\subsection{Effect of Number of Forward Pass}
\label{sec:num_pass}

\begin{table}
    \centering
        \resizebox{0.35\textwidth}{!}{
    \begin{tabular}{c||c|c|c|c|c} \hline
        num of pass	& 1 & 4 & 6 & 8	& 10 \\ \hline \hline
        AUC	& 0.9651 & 0.9892 & 0.9918 & 0.9908	& 0.9897\\ \hline
tpr(fpr@0.01) &	0.5471&	0.7333&	0.7891&	0.8051&	0.7801\\ \hline
tpr(fpr@0.05) &	0.8362&	0.9569&	0.9709&	0.9720&	0.9654\\ \hline
tpr(fpr@0.10) &	0.9013&	0.9990&	0.9994&	0.9903&	0.9865\\ \hline
    \end{tabular}
    }
    \caption{Effect of the number of forward passes on MNIST against PGD attack. {Increasing the number of passes is helpful but a very large number is not necessary as $4$ passes already shows reasonably good results.}}
    \label{tab:forward}
\end{table}

The proposed method is based on two blocks: 1) The first part is that the distributional difference between natural/adversarial images of BNN is larger compared to that of DNN. Unfortunately, we cannot prove this part theoretically, but observe the phenomenon empirically (e.g., Figure~\ref{fig:bd_dis}). 2) Proposition 1 shows that this distributional difference can be enlarged by leveraging the randomness of the BNN model (through multiple passes). Ideally, we need to generate distributions from an infinite number of passes, which is impossible in real practice. Therefore, we conducted experiments to study the effect of the number of forward passes on MNIST against PGD attack. The $\ell_\infty$ norm of PGD attack is bounded by 0.3 in the experiments. 

As shown in Table~\ref{tab:forward}, a few passes can recover this property. Comparing the performance of $4$ passes and $1$ pass, we see that increasing the number of passes helps improve performance. However, after a certain point, this increase does not improve the performance much. {Therefore, we do not need to worry that too many forward passes will be required for the distribution simulation.}

\vspace{-10pt}
\subsection{Defense against Attack with Different Parameters}
\label{sec:attack_param}

{Some previous works~\cite{athalye2018obfuscated} pointed out that detection methods can fail when the adversarial attacks are strong, such as C$\&$W attack with high confidence. 
Therefore, we test \textsc{BATer} against adversarial attacks of different strengths across three datasets. For PGD and FGSM attacks, the parameter $\epsilon$ captures the strength of the attack with larger $\epsilon$ representing a stronger attack. For C$\&$W, we try different confidence levels. 
The performance of \textsc{BATer} is reported in Table~\ref{tab:param}. Out of 27 AUC values, 24 of them are above 0.980 and all the AUCs are above 0.920. 
The results show that \textsc{BATer} performs well against various adversarial attacks with different strengths.}

\begin{table}
    \centering
    
\resizebox{0.5\textwidth}{!}{
    \begin{tabular}{c|c|c|c|c||c|c|c||c|c|c}
    \hline
    Data   & Metric/Parameter & \multicolumn{3}{c||}{C$\&$W} &  \multicolumn{3}{c||}{FGSM}  & \multicolumn{3}{c}{PGD}  \\ \hline \hline
    \multirow{5}{*}{CIFAR10}    & Parameter Value &  0   &  10 &  20 & 0.01 & 0.03 & 0.05 & 0.01 & 0.03 & 0.05   \\ \cline{2-11} 
            & AUC   & 0.980  &  0.999    &  0.995 & 0.982  &  0.995    &  0.996  & 0.965  &  0.971    &  0.981   \\ 
            & TPR(FPR@0.01) & 0.606  &  0.998   &  0.939 & 0.497  &  0.878  &  0.839  & 0.287  &  0.813  &  0.834  \\ 
            & TPR(FPR@0.05) &  0.881 &  1.000   &  0.995  &  0.942 &  0.991   &  0.996 &  0.917 &  0.881   &  0.928 \\ 
            & TPR(FPR@0.10) & 0.965  &  1.000   &  0.995 & 0.978  &  0.998   &  0.996 & 0.960  &  0.917   &  0.957 \\ \hline \hline
    \multirow{5}{*}{MNIST}     & Parameter Value & 0 & 10 & 20 & 0.1 & 0.3 & 0.5 & 0.1 & 0.3 & 0.5 \\ \cline{2-11}
            & AUC   & 0.999  &  0.995    &  0.995  & 0.993  & 0.999    &  0.999 & 0.980  & 0.989    &  0.996  \\ 
            & TPR(FPR@0.01) & 0.974 &  0.913   &  0.919 & 0.817 &  0.972   &  1.000 & 0.692 &  0.733   &  0.920  \\ 
            & TPR(FPR@0.05) &  0.997 &  0.993  &  0.994 &  0.994 &  0.998  &  1.000 &  0.973 &  0.957  &  0.992 \\ 
            & TPR(FPR@0.10) & 1.000 &  0.998   &  0.999  & 0.998 &  1.000   &  1.000 & 0.992 &  0.999   &  0.996 \\ \hline \hline
            & Parameter Value & 0 & 10 & 20 & 0.01 & 0.02 & 0.03 & 0.01 & 0.02 & 0.03\\ \cline{2-11} 
    \multirow{2}{*}{Imagenet}& AUC &  0.941  &  0.991 & 0.983  &  0.989  & 0.992  &  0.994  &  0.976  & 0.982  &  0.987  \\ 
               & TPR(FPR@0.01) & 0.146  &  0.896   &    0.642 & 0.569  &  0.824   &   0.841 & 0.729  &  0.511   &   0.708  \\ 
  \multirow{2}{*}{-sub}    & TPR(FPR@0.05) & 0.538  &  0.951   &    0.910 & 0.989  &   0.985  &  0.995 & 0.904  &  0.936  &  0.951  \\ 
               & TPR(FPR@0.10) & 0.815  &  0.977   &    0.964 & 1.000  &  0.997   &   0.999 & 0.947  &  0.980   &  0.984 \\ \hline
    \end{tabular}
    }
    \caption{ Performance of detection methods against adversarial attacks with different parameters. Out of 27 AUC values, 24 of them are above 0.980 and all the AUCs are above 0.920. \textsc{BATer} performs well against attacks of different strengths.}
    \label{tab:param}
    \vspace{-10pt}
\end{table}

\vspace{-10pt}
\subsection{Adaptive Attack}
\label{sec:custom_attack}

{All the previous experiments are carried out in a gray-box setting, where we assume the adversary has access to the classifier model but does not know the details of the detector. The white-box setting assumes that the adversary has access to both the classifier and the detector. Therefore, an adaptive attack method can be built to attack both the classifier and the detector. This is worth studying as it can reveal possible drawbacks of the method and promote future research direction. 

To develop an adaptive attack against \textsc{BATer}, we propose the following objective:
\begin{align}
\label{eq:white_box}
    \argmin\limits_{\|\bx-\bx_0\|_\infty\le\epsilon} -L_1(\bx,y_0)-\lambda L_2(\bx,z_0),
\end{align}
where $L_1$ and $L_2$ represent the classification loss and detection loss respectively, $\lambda$ controls the trade-off between the two, $y_0$ is the label of original input, $z_0$ is the detection label, and $\bx$ and $\bx_0$ represent adversarial example and original input. The loss function aims to fool the classifier and the detector at the same time. In the experiment, we set $\lambda=1$. To optimize over the loss function, we build a torch version of the Wasserstein distance function based on the one from the scipy package, making it possible to get the gradient of the second part of the loss function. Due to the sorting operations in the Wasserstein distance calculation, the function is non-differentiable at some points. However, if we are not at those points we can assume the permutation won't change within a small region, so it becomes differentiable using the same permutation forward and backward. So, the gradient is still an approximation but very close. }

\begin{table}[H]
    \centering
\resizebox{0.35\textwidth}{!}{
    \begin{tabular}{c|c|c|c}
    \hline
    Metric        &  MNIST & CIFAR10 & Imagenet-sub  \\ \hline \hline
    Robust.Acc    & 0.203 & 0.112 & 0.215 \\ \hline
    AUC           & 0.644 & 0.801 & 0.583   \\ \hline
    TPR(FPR@0.01) & 0.325 & 0.134 & 0.051  \\ \hline
    TPR(FPR@0.05) & 0.432 & 0.346 & 0.171 \\ \hline
    TPR(FPR@0.10) & 0.476 & 0.450 & 0.256 \\ \hline
    \end{tabular}
    }
    \caption{Peformance of \textsc{BATer} against adaptive attack. Considering both robust accuracy and detection AUC, \textsc{BATer} shows acceptable performance against the adaptive attack.}
    \label{tab:custom}
    \vspace{-10pt}
\end{table}

The performance of \textsc{BATer} against the adaptive attack on $1000$ randomly selected images of each dataset is shown in Table~\ref{tab:custom}. We employ the same attack parameters as outlined in Section~\ref{sec:sota}.
Compared to the gray-box setting, the performance drops, but still reasonable and better than without the detection system. The task of fooling the detection part makes the robust accuracy increase. On MNIST, the robust accuracy increases to $20.3\%$ and the AUC drops to $0.644$. Taking both robust accuracy and detection AUC into consideration, the framework can still handle a reasonable portion of adversarial examples correctly. On CIFAR10, though the robust accuracy only increases to $11.2\%$, the detection AUC is $0.801$. On Imagenet-sub, the performance is similar to that on MNIST. 

\section{Conclusion}
\label{sec:con}

In this paper, we introduce a new framework to detect adversarial examples with Bayesian Neural Network, by capturing the distributional differences of multiple hidden layer outputs between the natural and adversarial examples. We show that our detection framework outperforms other state-of-the-art methods in detecting adversarial examples generated by various kinds of attacks. {It also displays strong performance in detecting adversarial examples generated by various attack methods with different strengths and adversarial examples generated by an adaptive attack method.}


\bibliographystyle{IEEEtran}
\bibliography{main}

\begin{thebibliography}{10}
\providecommand{\url}[1]{#1}
\csname url@samestyle\endcsname
\providecommand{\newblock}{\relax}
\providecommand{\bibinfo}[2]{#2}
\providecommand{\BIBentrySTDinterwordspacing}{\spaceskip=0pt\relax}
\providecommand{\BIBentryALTinterwordstretchfactor}{4}
\providecommand{\BIBentryALTinterwordspacing}{\spaceskip=\fontdimen2\font plus
\BIBentryALTinterwordstretchfactor\fontdimen3\font minus
  \fontdimen4\font\relax}
\providecommand{\BIBforeignlanguage}[2]{{%
\expandafter\ifx\csname l@#1\endcsname\relax
\typeout{** WARNING: IEEEtran.bst: No hyphenation pattern has been}%
\typeout{** loaded for the language `#1'. Using the pattern for}%
\typeout{** the default language instead.}%
\else
\language=\csname l@#1\endcsname
\fi
#2}}
\providecommand{\BIBdecl}{\relax}
\BIBdecl

\bibitem{goodfellow2014explaining}
I.~Goodfellow, J.~Shlens, and C.~Szegedy, ``Explaining and harnessing
  adversarial examples,'' in \emph{International Conference on Learning
  Representations}, 2015.

\bibitem{szegedy2013intriguing}
C.~Szegedy, W.~Zaremba, I.~Sutskever, J.~Bruna, D.~Erhan, I.~Goodfellow, and
  R.~Fergus, ``Intriguing properties of neural networks,'' \emph{arXiv preprint
  arXiv:1312.6199}, 2013.

\bibitem{kuurkova2018artificial}
V.~Kurkova, Y.~Manolopoulos, B.~Hammer, L.~Iliadis, and I.~Maglogiannis,
  \emph{Artificial Neural Networks and Machine Learning--ICANN 2018: 27th
  International Conference on Artificial Neural Networks, Rhodes, Greece,
  October 4-7, 2018, Proceedings, Part III}.\hskip 1em plus 0.5em minus
  0.4em\relax Springer, 2018, vol. 11141.

\bibitem{yang2020towards}
P.~Yang, \emph{Towards Adversarial Robustness of Deep Neural Networks}.\hskip
  1em plus 0.5em minus 0.4em\relax University of California, Davis, 2020.

\bibitem{li2020robustness}
Y.~Li, \emph{On Robustness and Efficiency of Machine Learning Systems}.\hskip
  1em plus 0.5em minus 0.4em\relax University of California, Davis, 2020.

\bibitem{li2022review}
Y.~Li, M.~Cheng, C.-J. Hsieh, and T.~C. Lee, ``A review of adversarial attack
  and defense for classification methods,'' \emph{The American Statistician},
  vol.~76, no.~4, pp. 329--345, 2022.

\bibitem{yuan2019attack}
X.~Yuan, P.~He, Q.~Zhu, and X.~Li, ``Adversarial examples: Attacks and defenses
  for deep learning,'' \emph{IEEE Transactions on Neural Networks and Learning
  Systems}, vol.~30, no.~9, pp. 2805--2824, 2019.

\bibitem{zhang2020oppo}
J.~Zhang and C.~Li, ``Adversarial examples: Opportunities and challenges,''
  \emph{IEEE Transactions on Neural Networks and Learning Systems}, vol.~31,
  no.~7, pp. 2578--2593, 2020.

\bibitem{chan2021break}
A.~Chan, L.~Ma, F.~Juefei-Xu, Y.-S. Ong, X.~Xie, M.~Xue, and Y.~Liu, ``Breaking
  neural reasoning architectures with metamorphic relation-based adversarial
  examples,'' \emph{IEEE Transactions on Neural Networks and Learning Systems},
  pp. 1--7, 2021.

\bibitem{li2021review}
\BIBentryALTinterwordspacing
\textbf{Yao Li}, M.~Cheng, C.-J. Hsieh, and T.~C.~M. Lee, ``A review of
  adversarial attack and defense for classification methods,'' \emph{The
  American Statistician}, vol.~0, no.~ja, pp. 1--44, 2021. [Online]. Available:
  \url{https://doi.org/10.1080/00031305.2021.2006781}
\BIBentrySTDinterwordspacing

\bibitem{athalye2018obfuscated}
A.~Athalye, N.~Carlini, and D.~Wagner, ``Obfuscated gradients give a false
  sense of security: Circumventing defenses to adversarial examples,'' in
  \emph{International Conference on Machine Learning (ICML)}, 2018.

\bibitem{dhillon2018stochastic}
G.~S. Dhillon, K.~Azizzadenesheli, Z.~C. Lipton, J.~Bernstein, J.~Kossaifi,
  A.~Khanna, and A.~Anandkumar, ``Stochastic activation pruning for robust
  adversarial defense,'' \emph{arXiv preprint arXiv:1803.01442}, 2018.

\bibitem{ma2018characterizing}
X.~Ma, B.~Li, Y.~Wang, S.~M. Erfani, S.~Wijewickrema, G.~Schoenebeck, D.~Song,
  M.~E. Houle, and J.~Bailey, ``Characterizing adversarial subspaces using
  local intrinsic dimensionality,'' \emph{International Conference on Learning
  Representations}, 2018.

\bibitem{samangouei2018defense}
P.~Samangouei, M.~Kabkab, and R.~Chellappa, ``Defense-gan: Protecting
  classifiers against adversarial attacks using generative models,''
  \emph{arXiv preprint arXiv:1805.06605}, 2018.

\bibitem{song2017pixeldefend}
Y.~Song, T.~Kim, S.~Nowozin, S.~Ermon, and N.~Kushman, ``Pixeldefend:
  Leveraging generative models to understand and defend against adversarial
  examples,'' \emph{arXiv preprint arXiv:1710.10766}, 2017.

\bibitem{xie2017mitigating}
C.~Xie, J.~Wang, Z.~Zhang, Z.~Ren, and A.~Yuille, ``Mitigating adversarial
  effects through randomization,'' \emph{arXiv preprint arXiv:1711.01991},
  2017.

\bibitem{madry2017towards}
A.~Madry, A.~Makelov, L.~Schmidt, D.~Tsipras, and A.~Vladu, ``Towards deep
  learning models resistant to adversarial attacks,'' \emph{arXiv preprint
  arXiv:1706.06083}, 2017.

\bibitem{tanay2016boundary}
T.~Tanay and L.~Griffin, ``A boundary tilting persepective on the phenomenon of
  adversarial examples,'' \emph{arXiv preprint arXiv:1608.07690}, 2016.

\bibitem{feinman2017detecting}
R.~Feinman, R.~R. Curtin, S.~Shintre, and A.~B. Gardner, ``Detecting
  adversarial samples from artifacts,'' \emph{International Conference on
  Machine Learning}, 2017.

\bibitem{zheng2018robust}
Z.~Zheng and P.~Hong, ``Robust detection of adversarial attacks by modeling the
  intrinsic properties of deep neural networks,'' in \emph{Advances in Neural
  Information Processing Systems}, 2018, pp. 7913--7922.

\bibitem{pang2018towards}
T.~Pang, C.~Du, Y.~Dong, and J.~Zhu, ``Towards robust detection of adversarial
  examples,'' in \emph{Advances in Neural Information Processing Systems},
  2018, pp. 4584--4594.

\bibitem{tao2018attacks}
G.~Tao, S.~Ma, Y.~Liu, and X.~Zhang, ``Attacks meet interpretability:
  Attribute-steered detection of adversarial samples,'' in \emph{Advances in
  Neural Information Processing Systems}, 2018, pp. 7717--7728.

\bibitem{lee2018simple}
K.~Lee, K.~Lee, H.~Lee, and J.~Shin, ``A simple unified framework for detecting
  out-of-distribution samples and adversarial attacks,'' in \emph{Advances in
  Neural Information Processing Systems}, 2018, pp. 7167--7177.

\bibitem{yang2020ml}
P.~Yang, J.~Chen, C.-J. Hsieh, J.-L. Wang, and M.~Jordan, ``Ml-loo: Detecting
  adversarial examples with feature attribution,'' in \emph{Proceedings of the
  AAAI Conference on Artificial Intelligence}, vol.~34, 2020, pp. 6639--6647.

\bibitem{ding2022consensus}
X.~Ding, Y.~Cheng, Y.~Luo, Q.~Li, and P.~Gope, ``Consensus adversarial defense
  method based on augmented examples,'' \emph{IEEE Transactions on Industrial
  Informatics}, vol.~19, no.~1, pp. 984--994, 2022.

\bibitem{liu2017towards}
X.~Liu, M.~Cheng, H.~Zhang, and C.-J. Hsieh, ``Towards robust neural networks
  via random self-ensemble,'' \emph{arXiv preprint arXiv:1712.00673}, 2017.

\bibitem{cohen2019certified}
J.~M. Cohen, E.~Rosenfeld, and J.~Z. Kolter, ``Certified adversarial robustness
  via randomized smoothing,'' \emph{arXiv preprint arXiv:1902.02918}, 2019.

\bibitem{liu2018advbnn}
X.~Liu, Y.~Li, C.~Wu, and C.-J. Hsieh, ``Adv-{BNN}: Improved adversarial
  defense through robust bayesian neural network,'' in \emph{International
  Conference on Learning Representations}, 2019.

\bibitem{ye2018bal}
\BIBentryALTinterwordspacing
N.~Ye and Z.~Zhu, ``Bayesian adversarial learning,'' in \emph{Advances in
  Neural Information Processing Systems 31}, S.~Bengio, H.~Wallach,
  H.~Larochelle, K.~Grauman, N.~Cesa-Bianchi, and R.~Garnett, Eds.\hskip 1em
  plus 0.5em minus 0.4em\relax Curran Associates, Inc., 2018, pp. 6892--6901.
  [Online]. Available:
  \url{http://papers.nips.cc/paper/7921-bayesian-adversarial-learning.pdf}
\BIBentrySTDinterwordspacing

\bibitem{carbone2020bnn}
G.~Carbone, M.~Wicker, L.~Laurenti, A.~Patane\textquotesingle, L.~Bortolussi,
  and G.~Sanguinetti, ``Robustness of bayesian neural networks to
  gradient-based attacks,'' in \emph{Advances in Neural Information Processing
  Systems}, H.~Larochelle, M.~Ranzato, R.~Hadsell, M.~F. Balcan, and H.~Lin,
  Eds., vol.~33.\hskip 1em plus 0.5em minus 0.4em\relax Curran Associates,
  Inc., 2020, pp. 15\,602--15\,613.

\bibitem{yang2022mitigating}
C.-H.~H. Yang, Z.~Ahmed, Y.~Gu, J.~Szurley, R.~Ren, L.~Liu, A.~Stolcke, and
  I.~Bulyko, ``Mitigating closed-model adversarial examples with bayesian
  neural modeling for enhanced end-to-end speech recognition,'' in \emph{ICASSP
  2022-2022 IEEE International Conference on Acoustics, Speech and Signal
  Processing (ICASSP)}.\hskip 1em plus 0.5em minus 0.4em\relax IEEE, 2022, pp.
  6302--6306.

\bibitem{blundell2015weight}
C.~Blundell, J.~Cornebise, K.~Kavukcuoglu, and D.~Wierstra, ``Weight
  uncertainty in neural network,'' in \emph{International Conference on Machine
  Learning}, 2015, pp. 1613--1622.

\bibitem{yuan2019adversarial}
X.~Yuan, P.~He, Q.~Zhu, and X.~Li, ``Adversarial examples: Attacks and defenses
  for deep learning,'' \emph{IEEE Transactions on Neural Networks and Learning
  Systems}, 2019.

\bibitem{carlini2017adversarial}
N.~Carlini and D.~Wagner, ``Adversarial examples are not easily detected:
  Bypassing ten detection methods,'' in \emph{Proceedings of the 10th ACM
  Workshop on Artificial Intelligence and Security}.\hskip 1em plus 0.5em minus
  0.4em\relax ACM, 2017, pp. 3--14.

\bibitem{carlini2017towards}
------, ``Towards evaluating the robustness of neural networks,'' in
  \emph{Security and Privacy (SP), 2017 IEEE Symposium on}.\hskip 1em plus
  0.5em minus 0.4em\relax IEEE, 2017, pp. 39--57.

\bibitem{moosavi2017universal}
S.-M. Moosavi-Dezfooli, A.~Fawzi, O.~Fawzi, and P.~Frossard, ``Universal
  adversarial perturbations,'' \emph{arXiv preprint}, 2017.

\bibitem{che2020smgea}
Z.~Che, A.~Borji, G.~Zhai, S.~Ling, J.~Li, X.~Min, G.~Guo, and P.~L. Callet,
  ``Smgea: A new ensemble adversarial attack powered by long-term gradient
  memories,'' \emph{IEEE Transactions on Neural Networks and Learning Systems},
  pp. 1--15, 2020.

\bibitem{liang2021explore}
L.~Liang, X.~Hu, L.~Deng, Y.~Wu, G.~Li, Y.~Ding, P.~Li, and Y.~Xie, ``Exploring
  adversarial attack in spiking neural networks with spike-compatible
  gradient,'' \emph{IEEE Transactions on Neural Networks and Learning Systems},
  pp. 1--15, 2021.

\bibitem{zhao2019admm}
P.~Zhao, K.~Xu, S.~Liu, Y.~Wang, and X.~Lin, ``Admm attack: an enhanced
  adversarial attack for deep neural networks with undetectable distortions,''
  in \emph{Proceedings of the 24th Asia and South Pacific Design Automation
  Conference}, 2019, pp. 499--505.

\bibitem{chen2019adversarial}
T.~Chen, J.~Liu, Y.~Xiang, W.~Niu, E.~Tong, and Z.~Han, ``Adversarial attack
  and defense in reinforcement learning-from ai security view,''
  \emph{Cybersecurity}, vol.~2, pp. 1--22, 2019.

\bibitem{moosavi2016deepfool}
S.-M. Moosavi-Dezfooli, A.~Fawzi, and P.~Frossard, ``Deepfool: a simple and
  accurate method to fool deep neural networks,'' in \emph{Proceedings of the
  IEEE Conference on Computer Vision and Pattern Recognition}, 2016, pp.
  2574--2582.

\bibitem{chen2017ead}
P.-Y. Chen, Y.~Sharma, H.~Zhang, J.~Yi, and C.-J. Hsieh, ``Ead: elastic-net
  attacks to deep neural networks via adversarial examples,'' in \emph{AAAI},
  2018.

\bibitem{carlini2018evaluation}
N.~Carlini, \emph{Evaluation and design of robust neural network
  defenses}.\hskip 1em plus 0.5em minus 0.4em\relax University of California,
  Berkeley, 2018.

\bibitem{chen2018ead}
P.-Y. Chen, Y.~Sharma, H.~Zhang, J.~Yi, and C.-J. Hsieh, ``Ead: elastic-net
  attacks to deep neural networks via adversarial examples,'' in
  \emph{Proceedings of the AAAI conference on artificial intelligence},
  vol.~32, no.~1, 2018.

\bibitem{chen2017zoo}
P.-Y. Chen, H.~Zhang, Y.~Sharma, J.~Yi, and C.-J. Hsieh, ``Zoo: Zeroth order
  optimization based black-box attacks to deep neural networks without training
  substitute models,'' in \emph{Proceedings of the 10th ACM Workshop on
  Artificial Intelligence and Security}.\hskip 1em plus 0.5em minus 0.4em\relax
  ACM, 2017, pp. 15--26.

\bibitem{brendel2017decision}
W.~Brendel, J.~Rauber, and M.~Bethge, ``Decision-based adversarial attacks:
  Reliable attacks against black-box machine learning models,'' \emph{arXiv
  preprint arXiv:1712.04248}, 2017.

\bibitem{ilyas2018black}
A.~Ilyas, L.~Engstrom, A.~Athalye, and J.~Lin, ``Black-box adversarial attacks
  with limited queries and information,'' \emph{arXiv preprint
  arXiv:1804.08598}, 2018.

\bibitem{cheng2019sign}
M.~Cheng, S.~Singh, P.-Y. Chen, S.~Liu, and C.-J. Hsieh, ``Sign-opt: A
  query-efficient hard-label adversarial attack,'' \emph{arXiv preprint
  arXiv:1909.10773}, 2019.

\bibitem{yan2019subspace}
Z.~Yan, Y.~Guo, and C.~Zhang, ``Subspace attack: Exploiting promising subspaces
  for query-efficient black-box attacks,'' \emph{arXiv preprint
  arXiv:1906.04392}, 2019.

\bibitem{chen2019hopskipjumpattack}
J.~Chen, M.~I. Jordan, and M.~J. Wainwright, ``Hopskipjumpattack: A
  query-efficient decision-based attack,'' \emph{arXiv preprint
  arXiv:1904.02144}, 2019.

\bibitem{chen2019towards}
J.~Chen, \emph{Towards Interpretability and Robustness of Machine Learning
  Models}.\hskip 1em plus 0.5em minus 0.4em\relax University of California,
  Berkeley, 2019.

\bibitem{kurakin2016adversarial}
A.~Kurakin, I.~Goodfellow, and S.~Bengio, ``Adversarial examples in the
  physical world,'' \emph{arXiv preprint arXiv:1607.02533}, 2016.

\bibitem{tramer2017ensemble}
F.~Tram{\`e}r, A.~Kurakin, N.~Papernot, I.~Goodfellow, D.~Boneh, and
  P.~McDaniel, ``Ensemble adversarial training: Attacks and defenses,''
  \emph{arXiv preprint arXiv:1705.07204}, 2017.

\bibitem{zhang2019theoretically}
H.~Zhang, Y.~Yu, J.~Jiao, E.~P. Xing, L.~E. Ghaoui, and M.~I. Jordan,
  ``Theoretically principled trade-off between robustness and accuracy,''
  \emph{arXiv preprint arXiv:1901.08573}, 2019.

\bibitem{ye2021acce}
N.~Ye, Q.~Li, X.-Y. Zhou, and Z.~Zhu, ``An annealing mechanism for adversarial
  training acceleration,'' \emph{IEEE Transactions on Neural Networks and
  Learning Systems}, pp. 1--12, 2021.

\bibitem{meng2017magnet}
D.~Meng and H.~Chen, ``Magnet: a two-pronged defense against adversarial
  examples,'' in \emph{Proceedings of the 2017 ACM SIGSAC Conference on
  Computer and Communications Security}.\hskip 1em plus 0.5em minus 0.4em\relax
  ACM, 2017, pp. 135--147.

\bibitem{li2020optimal}
Y.~Li, M.~R. Min, W.~Yu, C.-J. Hsieh, T.~Lee, and E.~Kruus, ``Optimal transport
  classifier: Defending against adversarial attacks by regularized deep
  embedding,'' \emph{arXiv preprint arXiv:1811.07950}, 2020.

\bibitem{jalal2017robust}
A.~Jalal, A.~Ilyas, C.~Daskalakis, and A.~G. Dimakis, ``The robust manifold
  defense: Adversarial training using generative models,'' \emph{arXiv preprint
  arXiv:1712.09196}, 2017.

\bibitem{li2021towards}
Y.~Li, M.~R. Min, T.~Lee, W.~Yu, E.~Kruus, W.~Wang, and C.-J. Hsieh, ``Towards
  robustness of deep neural networks via regularization,'' in \emph{Proceedings
  of the IEEE/CVF International Conference on Computer Vision}, 2021, pp.
  7496--7505.

\bibitem{wong2018provable}
E.~Wong and Z.~Kolter, ``Provable defenses against adversarial examples via the
  convex outer adversarial polytope,'' in \emph{International Conference on
  Machine Learning}.\hskip 1em plus 0.5em minus 0.4em\relax PMLR, 2018, pp.
  5286--5295.

\bibitem{everett2021cert}
M.~Everett, B.~Lütjens, and J.~P. How, ``Certifiable robustness to adversarial
  state uncertainty in deep reinforcement learning,'' \emph{IEEE Transactions
  on Neural Networks and Learning Systems}, pp. 1--15, 2021.

\bibitem{chen2021feature}
X.~Chen, J.~Weng, X.~Deng, W.~Luo, Y.~Lan, and Q.~Tian, ``Feature distillation
  in deep attention network against adversarial examples,'' \emph{IEEE
  Transactions on Neural Networks and Learning Systems}, pp. 1--15, 2021.

\bibitem{liu2021model}
Q.~Liu and W.~Wen, ``Model compression hardens deep neural networks: A new
  perspective to prevent adversarial attacks,'' \emph{IEEE Transactions on
  Neural Networks and Learning Systems}, pp. 1--12, 2021.

\bibitem{zhang2020challenging}
B.~Zhang, B.~Tondi, X.~Lv, and M.~Barni, ``Challenging the adversarial
  robustness of dnns based on error-correcting output codes,'' \emph{Security
  and Communication Networks}, vol. 2020, pp. 1--11, 2020.

\bibitem{mustafa2020deeply}
A.~Mustafa, S.~H. Khan, M.~Hayat, R.~Goecke, J.~Shen, and L.~Shao, ``Deeply
  supervised discriminative learning for adversarial defense,'' \emph{IEEE
  transactions on pattern analysis and machine intelligence}, vol.~43, no.~9,
  pp. 3154--3166, 2020.

\bibitem{zhao2022enhanced}
S.~Zhao, J.~Yu, Z.~Sun, B.~Zhang, and X.~Wei, ``Enhanced accuracy and
  robustness via multi-teacher adversarial distillation,'' in \emph{Computer
  Vision--ECCV 2022: 17th European Conference, Tel Aviv, Israel, October
  23--27, 2022, Proceedings, Part IV}.\hskip 1em plus 0.5em minus 0.4em\relax
  Springer, 2022, pp. 585--602.

\bibitem{metzen2017on}
\BIBentryALTinterwordspacing
J.~H. Metzen, T.~Genewein, V.~Fischer, and B.~Bischoff, ``On detecting
  adversarial perturbations,'' in \emph{International Conference on Learning
  Representations}, 2017. [Online]. Available:
  \url{https://openreview.net/forum?id=SJzCSf9xg}
\BIBentrySTDinterwordspacing

\bibitem{metzen2017detecting}
------, ``On detecting adversarial perturbations,'' \emph{arXiv preprint
  arXiv:1702.04267}, 2017.

\bibitem{agarwal2021damad}
A.~Agarwal, G.~Goswami, M.~Vatsa, R.~Singh, and N.~K. Ratha, ``Damad: Database,
  attack, and model agnostic adversarial perturbation detector,'' \emph{IEEE
  Transactions on Neural Networks and Learning Systems}, pp. 1--13, 2021.

\bibitem{nesti2021detect}
F.~Nesti, A.~Biondi, and G.~Buttazzo, ``Detecting adversarial examples by input
  transformations, defense perturbations, and voting,'' \emph{IEEE Transactions
  on Neural Networks and Learning Systems}, pp. 1--13, 2021.

\bibitem{gong2017adversarial}
Z.~Gong, W.~Wang, and W.-S. Ku, ``Adversarial and clean data are not twins,''
  \emph{arXiv preprint arXiv:1704.04960}, 2017.

\bibitem{sperl2020dla}
P.~Sperl, C.-Y. Kao, P.~Chen, X.~Lei, and K.~B{\"o}ttinger, ``Dla:
  dense-layer-analysis for adversarial example detection,'' in \emph{2020 IEEE
  European Symposium on Security and Privacy (EuroS\&P)}.\hskip 1em plus 0.5em
  minus 0.4em\relax IEEE, 2020, pp. 198--215.

\bibitem{gao2023detecting}
S.~Gao, R.~Wang, X.~Wang, S.~Yu, Y.~Dong, S.~Yao, and W.~Zhou, ``Detecting
  adversarial examples on deep neural networks with mutual information neural
  estimation,'' \emph{IEEE Transactions on Dependable and Secure Computing},
  2023.

\bibitem{chen2022adversarial}
Y.~Chen, M.~Zhang, J.~Li, and X.~Kuang, ``Adversarial attacks and defenses in
  image classification: A practical perspective,'' in \emph{2022 7th
  International Conference on Image, Vision and Computing (ICIVC)}.\hskip 1em
  plus 0.5em minus 0.4em\relax IEEE, 2022, pp. 424--430.

\bibitem{roth2019odds}
K.~Roth, Y.~Kilcher, and T.~Hofmann, ``The odds are odd: A statistical test for
  detecting adversarial examples,'' in \emph{International Conference on
  Machine Learning}.\hskip 1em plus 0.5em minus 0.4em\relax PMLR, 2019, pp.
  5498--5507.

\bibitem{raghuram2020detecting}
J.~Raghuram, V.~Chandrasekaran, S.~Jha, and S.~Banerjee, ``A general framework
  for detecting anomalous inputs to dnn classifiers,'' in \emph{International
  Conference on Machine Learning}.\hskip 1em plus 0.5em minus 0.4em\relax PMLR,
  2021, pp. 8764--8775.

\bibitem{gao2023towards}
Y.~Gao, Z.~Lin, Y.~Yang, and J.~Sang, ``Towards black-box adversarial example
  detection: A data reconstruction-based method,'' \emph{arXiv preprint
  arXiv:2306.02021}, 2023.

\bibitem{zhou-etal-2019-learning}
Y.~Zhou, J.-Y. Jiang, K.-W. Chang, and W.~Wang, ``Learning to discriminate
  perturbations for blocking adversarial attacks in text classification,'' in
  \emph{Proceedings of the 2019 Conference on Empirical Methods in Natural
  Language Processing and the 9th International Joint Conference on Natural
  Language Processing (EMNLP-IJCNLP)}, Nov. 2019.

\bibitem{mozes-etal-2021-frequency}
\BIBentryALTinterwordspacing
M.~Mozes, P.~Stenetorp, B.~Kleinberg, and L.~Griffin, ``Frequency-guided word
  substitutions for detecting textual adversarial examples,'' in
  \emph{Proceedings of the 16th Conference of the European Chapter of the
  Association for Computational Linguistics: Main Volume}.\hskip 1em plus 0.5em
  minus 0.4em\relax Online: Association for Computational Linguistics, Apr.
  2021, pp. 171--186. [Online]. Available:
  \url{https://aclanthology.org/2021.eacl-main.13}
\BIBentrySTDinterwordspacing

\bibitem{yoo-etal-2022-detection}
\BIBentryALTinterwordspacing
K.~Yoo, J.~Kim, J.~Jang, and N.~Kwak, ``Detection of adversarial examples in
  text classification: Benchmark and baseline via robust density estimation,''
  in \emph{Findings of the Association for Computational Linguistics: ACL
  2022}.\hskip 1em plus 0.5em minus 0.4em\relax Dublin, Ireland: Association
  for Computational Linguistics, May 2022, pp. 3656--3672. [Online]. Available:
  \url{https://aclanthology.org/2022.findings-acl.289}
\BIBentrySTDinterwordspacing

\bibitem{yin2022addmu}
F.~Yin, \textbf{Yao Li}, C.-J. Hsieh, and K.-W. Chang, ``Addmu: Detection of
  far-boundary adversarial examples with data and model uncertainty
  estimation,'' in \emph{Proceedings of the 2022 Conference on Empirical
  Methods in Natural Language Processing (EMNLP) (acceptance ratio 20.8\%)},
  December 2022.

\bibitem{ren2021adversarial}
H.~Ren, T.~Huang, and H.~Yan, ``Adversarial examples: attacks and defenses in
  the physical world,'' \emph{International Journal of Machine Learning and
  Cybernetics}, pp. 1--12, 2021.

\bibitem{simonyan2014very}
K.~Simonyan and A.~Zisserman, ``Very deep convolutional networks for
  large-scale image recognition,'' \emph{arXiv preprint arXiv:1409.1556}, 2014.

\bibitem{lecun1998mnist}
Y.~LeCun, ``The mnist database of handwritten digits,'' \emph{http://yann.
  lecun. com/exdb/mnist/}, 1998.

\bibitem{krizhevsky2009learning}
A.~Krizhevsky and G.~Hinton, ``Learning multiple layers of features from tiny
  images,'' Citeseer, Tech. Rep., 2009.

\bibitem{miyato2018spectral}
T.~Miyato, T.~Kataoka, M.~Koyama, and Y.~Yoshida, ``Spectral normalization for
  generative adversarial networks,'' \emph{arXiv preprint arXiv:1802.05957},
  2018.

\bibitem{zhang2018noisy}
G.~Zhang, S.~Sun, D.~Duvenaud, and R.~Grosse, ``Noisy natural gradient as
  variational inference,'' in \emph{International Conference on Machine
  Learning}.\hskip 1em plus 0.5em minus 0.4em\relax PMLR, 2018, pp. 5852--5861.

\bibitem{rauber2017foolbox}
J.~Rauber, W.~Brendel, and M.~Bethge, ``Foolbox: A python toolbox to benchmark
  the robustness of machine learning models,'' \emph{arXiv preprint
  arXiv:1707.04131}, 2017.

\bibitem{papernot2018deep}
N.~Papernot and P.~McDaniel, ``Deep k-nearest neighbors: Towards confident,
  interpretable and robust deep learning,'' \emph{arXiv preprint
  arXiv:1803.04765}, 2018.

\bibitem{jiang2018trust}
H.~Jiang, B.~Kim, M.~Y. Guan, and M.~R. Gupta, ``To trust or not to trust a
  classifier.'' in \emph{NeurIPS}, 2018, pp. 5546--5557.

\end{thebibliography}

\begin{IEEEbiography}[{\includegraphics[width=1in,height=1.25in,clip,keepaspectratio]{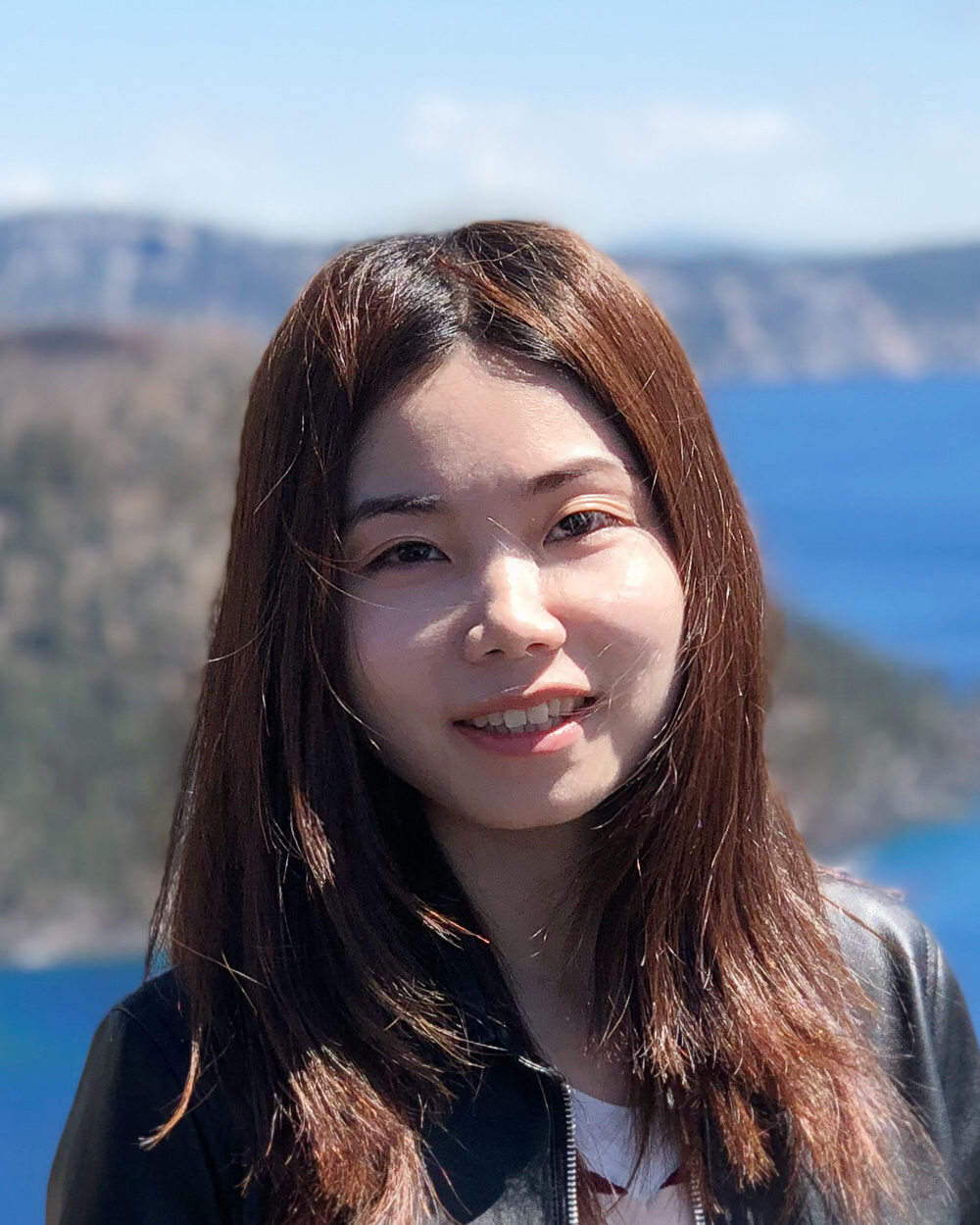}}]{Yao Li}
received the bachelor's (Statistics) degree in 2014 at Fudan University. In 2020, she completed a Ph.D. degree at the University of California, Davis. Currently, she is an assistant professor of Statistics and Operations Research at the University of North Carolina at Chapel Hill. Her research interests include trustworthy machine learning, computational pathology, and machine learning applications in other scientific disciplines.
\end{IEEEbiography}

\begin{IEEEbiography}[{\includegraphics[width=1in,height=1.25in,clip,keepaspectratio]{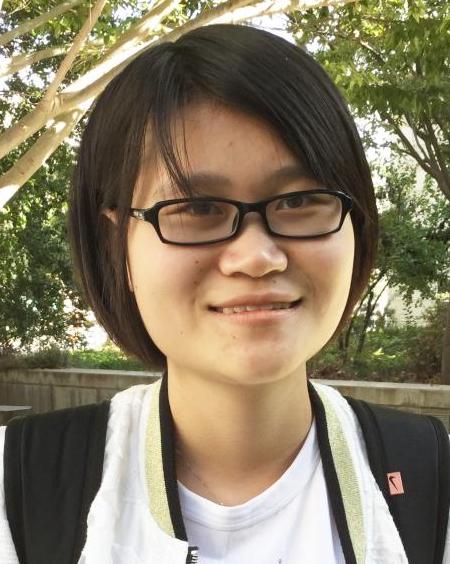}}]{Tongyi Tang}
received the bachelor's (Math) degree in 2016 at Fudan University. In 2021, she completed a Ph.D. degree at the University of California, Davis. Currently, she is a research scientist at Meta Platforms, Inc. Her research interests include optimization, random vector field modeling, and security of deep learning models.
\end{IEEEbiography}

\begin{IEEEbiography}[{\includegraphics[width=1in,height=1.25in,clip,keepaspectratio]{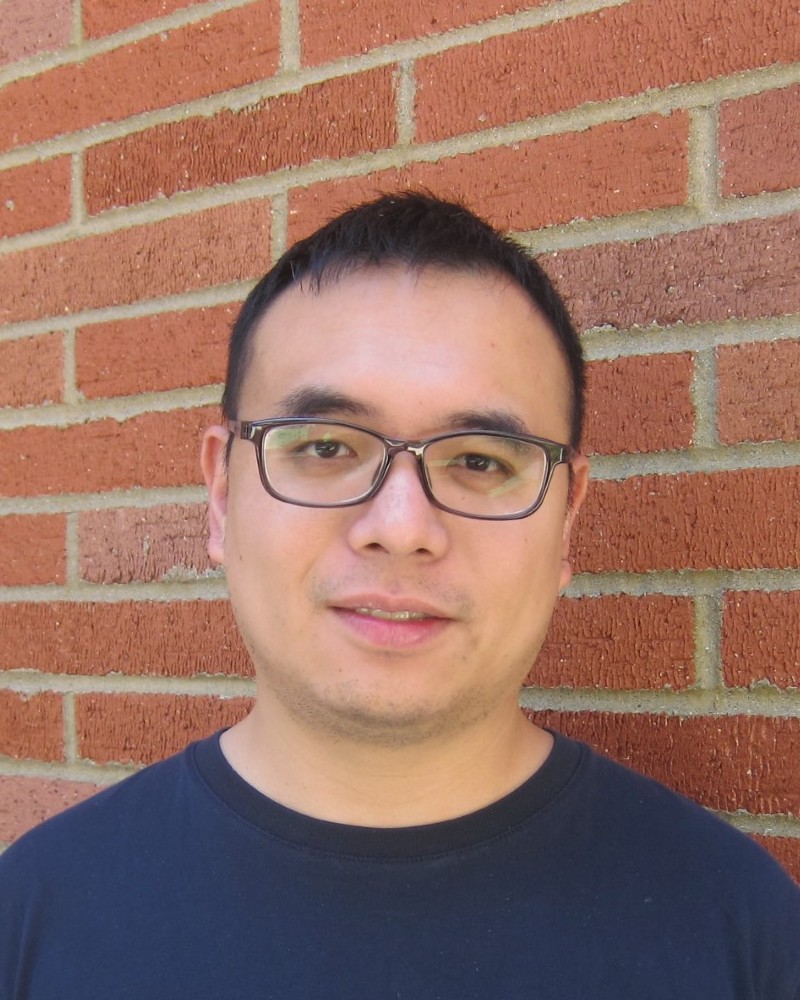}}]{Cho-Jui Hsieh} is an associate professor in the Computer Science Department at UCLA. His work primarily focuses on enhancing the efficiency and robustness of machine learning systems, and he has made significant contributions to multiple widely-used machine learning packages. He has been honored with the NSF Career Award, Samsung AI Researcher of the Year, and Google Research Scholar Award, and his work has been acknowledged with several paper awards in ICLR, KDD, ICDM, ICPP, and SC.
\end{IEEEbiography}

\begin{IEEEbiography}[{\includegraphics[width=1in,height=1.25in,clip,keepaspectratio]{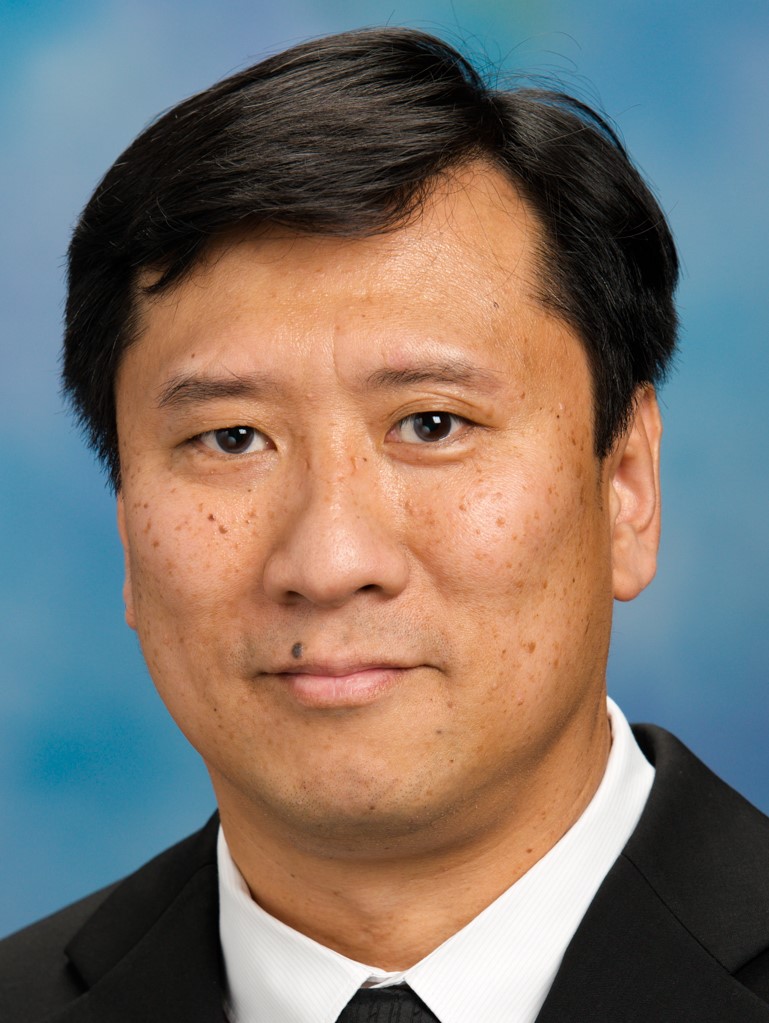}}]{Thomas C. M. Lee}
received the B.App.Sc. (Math) degree in 1992, and the B.Sc. (Hons) (Math) degree with University Medal in 1993, all from the University of Technology, Sydney, Australia. In 1997, he completed a Ph.D. degree jointly at Macquarie University and CSIRO Mathematical and Information Sciences, Sydney, Australia.

Currently, he is Professor of Statistics and Associate Dean of the Faculty of Mathematical and Physical Sciences at the University of California, Davis. He is an elected Fellow of the American Association for the Advancement of Science (AAAS), the American Statistical Association (ASA), and the Institute of Mathematical Statistics (IMS). From 2013 to 2015, he served as the Editor-in-Chief for the {\it Journal of Computational and Graphical Statistics}, from 2015 to 2018, he served as the Chair of the Department of Statistics at UC Davis, and currently he is the Review Editor for the {\it Journal of the American Statistical Association}. His research interests include inference methods, machine learning, and statistical applications in other scientific disciplines.
\end{IEEEbiography}

\newpage

\appendix

\subsection{Datasets and DNN Architectures}
\label{sec:data_arc}

One GPU (RTX 2080 Ti) is used in the experiment. The datasets used in the experiment do not contain personally identifiable information or offensive content. A summary of the datasets used, DNN architectures and test set performance of the corresponding DNNs are given in Table~\ref{tab:data}. For BNN, the structures used on the datasets are the same, except that the weights of BNN are not deterministic but follow Gaussian distributions. See a summary of BNN architectures and test set performance in Table~\ref{tab:bnn_arch}.

\begin{table}[H]
    \centering
    \caption{ Datasets and DNN Architectures. Conv.: Convolutional layer, FC: Fully Connected layer, BN: Batch Normalization.}
    \label{tab:data}
    \resizebox{0.48\textwidth}{!}{
    \begin{tabular}{c|c|c|c}
    \hline
    Dataset      & Number of Classes & Test Accuracy & Architecture \\ \hline\hline
    MNIST~\cite{lecun1998mnist}        & 10  & 99.20 & 2Conv. + 2FC layers   \\
    CIFAR10~\cite{krizhevsky2009learning}      & 10  & 93.34 & VGG16 with BN~\cite{simonyan2014very} \\
    Imagenet-Sub~\cite{miyato2018spectral} & 143 & 65.05 & VGG16 with BN~\cite{simonyan2014very}   \\ \hline
    \end{tabular}
    }
\end{table}

\begin{table}[h]
    \centering
    \caption{ Datasets and BNN Architectures. Conv.: Convolutional layer, FC: Fully Connected layer, BN: Batch Normalization. The structures of BNN are the same as DNN except that the weights of BNN are not deterministic but follow Gaussian distributions.}
    \label{tab:bnn_arch}
    \resizebox{0.48\textwidth}{!}{
    \begin{tabular}{c|c|c|c}
    \hline
    Dataset      & Number of Classes & Test Accuracy & Architecture \\ \hline\hline
    MNIST~\cite{lecun1998mnist}        & 10  & 98.81 & 2Conv. + 2FC layers   \\
    CIFAR10~\cite{krizhevsky2009learning}      & 10  & 92.32 & VGG16 with BN~\cite{simonyan2014very} \\
    Imagenet-Sub~\cite{miyato2018spectral} & 143 & 61.50  & VGG16 with BN~\cite{simonyan2014very}   \\ \hline
    \end{tabular}
    }
\end{table}

\subsection{Parameters of Attack Methods}
\label{sec:para}

Foolbox~\cite{rauber2017foolbox} is used to generate adversarial examples with C$\&$W, FGSM and PGD. The parameters used for the attack methods in Section~\ref{sec:sota}, \ref{sec:ablation}, \ref{sec:transfer}, and~\ref{sec:num_pass} are listed below:
\begin{itemize}
    \item C$\&$W~\cite{carlini2017towards} with $\ell_2$ norm: the confidence is set to $0$ for all three datasets except in high-confidence experiments, where the confidence values are discussed in section~\ref{sec:custom_attack}. The maximum number of iterations is set to $100$.
    \item FGSM~\cite{goodfellow2014explaining} with $\ell_\infty$ norm: the maximum $\epsilon$ values of FGSM attack on MNIST, CIFAR10 and Imagenet-sub are set to $0.3$, $0.03$ and $0.01$ respectively.
    \item PGD~\cite{madry2017towards} with $\ell_\infty$ norm: the $\epsilon$ values of PGD attack on MNIST, CIFAR10 and Imagenet-sub are set to $0.3$, $0.03$ and $0.01$ respectively. The number of iterations on MNIST, CIFAR10 and Imagenet-sub are set to $40$, $10$ and $10$, respectively. The step sizes of PGD attack on MNIST, CIFAR10 and Imagenet-sub are set to $0.1$, $0.01$ and $0.001$, respectively. 
\end{itemize}

For the adaptive attack, the parameter values are the same as PGD on three datasets. 

\subsection{Layer Selection}
\label{sec:layer}

\begin{table}
    \centering
    \caption{ Selected layers and test statistics used in the experiments}
    \label{tab:layer_selection}
    \resizebox{0.48\textwidth}{!}{
    \begin{tabular}{c|c|c|c|c|c|c}
    \hline
    \multirow{2}{*}{Data}     & \multicolumn{2}{c|}{C$\&$W} &  \multicolumn{2}{c|}{FGSM} &  \multicolumn{2}{c}{PGD} \\\cline{2-7}
                              & Layers  &  Statistic       & Layers  &  Statistic      & Layers  &  Statistic   \\
    \hline\hline
    MNIST        & [3, 5, 6]  & min & [3, 5, 6] & min  & [3, 5, 6] & min  \\
    CIFAR10      & [4, 7]   & min & [7] & mean &  [39, 42, 43] & mean\\
    Imagenet-Sub & [3, 7] & min & [7] & min & [39, 42, 43]  & mean \\ \hline
    \end{tabular}
    }
\end{table}

A summary of the selected layers is given in Table~\ref{tab:layer_selection}. 
Cross-validation is performed to perform layer selection by fitting a binary classifier (logistic regression) with each single layer's dispersion score. Layers with top-ranked performance measured by AUC scores are selected. In general, the top three layers will be selected, but if the top two or top one can yield similar performance, fewer number of layers will be selected for simplicity.
When performing a transfer attack, to align the input dimension of the attack detector, we select the same layers across different attacks. The layers selected are $[3,5,6]$, $[3,4,7,39,42,43]$, and $[39,42,43]$ on MNIST, CIFAR10 and Imagenet-sub respectively. The distribution of each hidden layer in BNN is simulated with 4 forward passes of the inputs. We also list the summary statistics we use as a measure of the average distance from the testing sample to the natural image sets.

\subsection{Implementation of Detection Methods}
\label{sec:impl}

Implementation details of detection methods KD, LID, ODD, JTLA and \textsc{BATer} are discussed in the following part:
\begin{itemize}
    \item KD~\cite{feinman2017detecting}: The Kernel Density Detection method is implemented by converting the authors' implementation into pytorch version. The author implementation is available at Github~\footnote{
\url{https://github.com/rfeinman/detecting-adversarial-samples}}. The default parameter values are used for the experiments in this paper.
    \item LID~\cite{ma2018characterizing}: The Local Intrinsic Dimension detection is implemented by using the code from Github~\footnote{\url{https://github.com/pokaxpoka/deep_Mahalanobis_detector}}. The numbers of neighbors used to calculate local intrinsic dimension are $10, 20, 30, 40, 50, 60, 70, 80$ and $90$.
    \item ODD~\cite{roth2019odds}: Odds are odd is implemented using the authors' original implementation with code available at Github~\footnote{\url{https://github.com/yk/icml19_public}}. The default parameter values are used for the experiment in this paper.
    \item JTLA~\cite{raghuram2020detecting}: Joint statistical Testing across DNN Layers for Anomalies is implemented using the authors' original implementation with code available at Github~\footnote{\url{https://github.com/jayaram-r/adversarial-detection}}. The test statistic is multinomial. The scoring method is p-value. The Fisher method is used to combine the p-values from different layers. For other parameters, the default values are used.
    \item \textsc{BATer}: Code is available at Github~\footnote{\url{https://github.com/BayesianDetection/BayesianDetection}}.
\end{itemize}

\subsection{Examples of Detected Adversarial Images}

\begin{figure}
    \centering
    \includegraphics[width=0.15\textwidth]{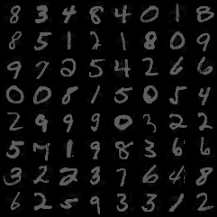}
    \includegraphics[width=0.15\textwidth]{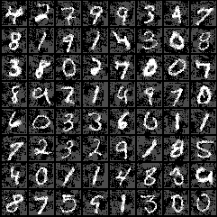}
    \includegraphics[width=0.15\textwidth]{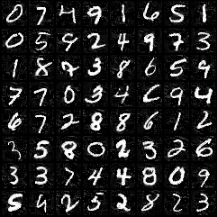}
    \includegraphics[width=0.15\textwidth]{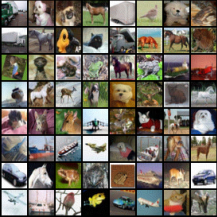}
    \includegraphics[width=0.15\textwidth]{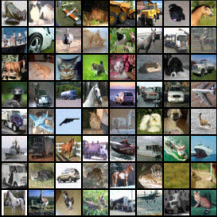}
    \includegraphics[width=0.15\textwidth]{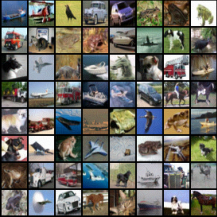}
    \includegraphics[width=0.15\textwidth]{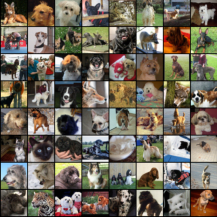}
    \includegraphics[width=0.15\textwidth]{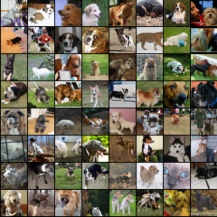}
    \includegraphics[width=0.15\textwidth]{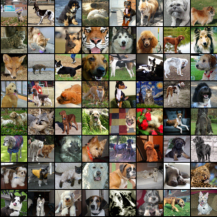}
    \caption{Examples of detected adversarial images. The three columns of adversarial images are generated from C\&W, FGSM and PGD attacks, respectively.}
    \label{fig:detected_adv}
\end{figure}

{In this part, we show examples of detected adversarial images in Figure~\ref{fig:detected_adv}. There are three columns in Figure~\ref{fig:detected_adv}. From left to right, the images in the columns are generated from C\&W, FGSM and PGD attacks, respectively. The corresponding attack parameters are the same as what we used in the main experiments. See details in Appendix~\ref{sec:para}.}

\subsection{Theoretical Analysis}
\label{sec:theo}

To show the advantage of BNN over DNN, there are two aspects. 1) BNN can better separate the distribution of natural hidden output and the distribution of natural hidden output. Figure~\ref{fig:bd_dis} empirically shows it. 2) Randomness/Variance involved in the parameters can help enlarge the distributional differences between natural and adversarial outputs. The following analysis shows the second aspect.

\subsubsection{Wasserstein Distance}
In the paper, we estimate the 1-Wasserstein distance between the distributions to detect adversarial examples. Here, we first show that BNN can enlarge distributional differences with this distance metric. 

For a model $f(\bx, \bw)$ with $\bx\sim \boldsymbol{D}_\bx$ and $\bw \sim \boldsymbol{D}_\bw$, where $\boldsymbol{D}_\bw$ is any distribution that satisfies $\bw$ is symmetric about $\bw_0 = \mathbb{E}[\bw]$, such as $\mathcal{N}(\bw_0,\boldsymbol{I})$. We want to show that 
\begin{align*}
\mathcal{W}_1(f(\bx+\vdelta, \bw), f(\bx, \bw)) \geq \mathcal{W}_1(f(\bx+\vdelta, \bw_0), f(\bx, \bw_0)),
\end{align*}
where $\mathcal{W}_1$ represents the 1-Wasserstein distance, which measures distance between distribution of $f(\bx+\vdelta, \bw)$ and distribution of $f(\bx, \bw)$.

As $f(\bx + \vdelta, \bw) \approx f(\bx, \bw) + \vdelta^T\nabla_\bx f(\bx, \bw)$, the distance can be approximated by
\begin{align*}
&\mathcal{W}_1(f(\bx+\vdelta, \bw_0), f(\bx, \bw_0)) \\
\approx & \vdelta \inf_{\pi}\int \|\vdelta^T\nabla_\bx f(\bx, \bw_0)\|\pi(\bx), \\
&\mathcal{W}_1(f(\bx+\vdelta, \bw), f(\bx, \bw)) \\
\approx &\inf_{\pi\in\Gamma(\mathcal{D}_\bx, \mathcal{D}_\bw)}\int \|\delta^T\nabla_\bx f(\bx, \bw)\|\pi(\bx, \bw).
\end{align*}

If $\nabla_\bx f(\bx, \bw)$ can be further simplified with $\nabla_\bx f(\bx, \bw_0) + \nabla_\bw \nabla_\bx f(\bx, \bw_0)  (\bw - \bw_0)$, 
\begin{align*}
&\mathcal{W}_1(f(\bx+\vdelta, \bw), f(\bx, \bw)) \\
~\approx &\inf_{\pi\in\Gamma(\mathcal{D}_\bx, \mathcal{D}_\bw)}\int \|\vdelta^T\nabla_\bx f(\bx, \bw_0) \\
&+ \vdelta^T \nabla_\bw \nabla_\bx f(\bx, \bw_0)  (\bw - \bw_0)\|\pi(\bx, \bw)   \\
~ =&  \inf_{\pi\in\Gamma(\mathcal{D}_\bx, \mathcal{D}_\bw)}\int \|\vdelta^T\nabla_\bx f(\bx, \bw_0) \\
&+ \vdelta^T\nabla_\bw \nabla_\bx f(\bx, \bw_0) (\bw_0 - \bw) \|\pi(\bx, \bw)   \\
~ \geq & \inf_{\pi}\int \|\vdelta^T\nabla_\bx f(\bx, \bw_0)\|\pi(\bx) \\
~ =& \mathcal{W}_1(f(\bx+\vdelta, \bw_0), f(\bx, \bw_0))
\end{align*}

\subsubsection{General Distance}
The inequality can be extended to any translation-invariant distance. For a model $f(\bx, \bw)$, where $\bx\sim \boldsymbol{D}_\bx$ and $\bw \sim \boldsymbol{D}_\bw$, then we want to show that
\begin{align*}
 \mathcal{D}(f(\bx+\vdelta, \bw), f(\bx, \bw)) \geq \mathcal{D}(f(\bx+\vdelta, \bw_0), f(\bx, \bw_0)),   
\end{align*}
where $\vdelta$ represents adversarial perturbation and $\mathcal{D}$ represents a translation-invariant distance measuring distribution dispersion.

As $f(\bx + \vdelta, \bw) \approx f(\bx, \bw) + \vdelta^T\nabla_\bx f(\bx, \bw)$ when $\|\vdelta\|$ is small, the distance can be approximated by
\begin{align*}
    &\mathcal{D}(f(\bx+\vdelta, \bw_0), f(\bx, \bw_0)) \\
    \approx &\mathcal{D}( f(\bx, \bw_0) + \vdelta^T\nabla_\bx f(\bx, \bw_0), f(\bx, \bw_0)) \\
    =& \mathcal{D}(\vdelta^T\nabla_\bx f(\bx, \bw_0), \boldsymbol{0}),\\
 &\mathcal{D}(f(\bx+\vdelta, \bw), f(\bx, \bw)) \approx \mathcal{D}(\vdelta^T\nabla_\bx f(\bx, \bw), \boldsymbol{0})   
\end{align*}

If $\nabla_\bx f(\bx, \bw)$ can be further simplified with $\nabla_\bx f(\bx, \bw_0) + \nabla_\bw \nabla_\bx f(\bx, \bw_0)  (\bw - \bw_0)$, 
\begin{align*}
&\mathcal{D}(f(\bx+\vdelta, \bw), f(\bx, \bw)) \\
~\approx &\mathcal{D}(\vdelta^T\nabla_\bx f(\bx, \bw_0) + \vdelta^T \nabla_\bw \nabla_\bx f(\bx, \bw_0)  (\bw - \bw_0), \boldsymbol{0})   \\
~ =&  \mathcal{D}(\vdelta^T\nabla_\bx f(\bx, \bw_0), \vdelta^T \nabla_\bw \nabla_\bx f(\bx, \bw_0)  (\bw_0 - \bw))   \\
~ =&  \mathcal{D}(-\vdelta^T\nabla_\bx f(\bx, \bw_0), \vdelta^T \nabla_\bw \nabla_\bx f(\bx, \bw_0)  (\bw_0 - \bw))   \\
~ \geq & \frac{1}{2}\mathcal{D}(\vdelta^T\nabla_\bx f(\bx, \bw_0), -\vdelta^T\nabla_\bx f(\bx, \bw_0))   \\
~ =& \frac{1}{2}\mathcal{D}(\vdelta^T\nabla_\bx f(\bx, \bw_0), \boldsymbol{0})   \\
~ \approx &\mathcal{D}(f(\bx+\vdelta, \bw_0), f(\bx, \bw_0))
\end{align*}
where the third equality derives from the translation-invariant of the metric and $\bw_0 - \bw \sim \bw - \bw_0$ in distribution.

\vfill

\end{document}